%% file: main.tex
\documentclass[11pt]{article}

\usepackage[preprint]{acl}

\usepackage{times}
\usepackage{latexsym}

\usepackage[T1]{fontenc}

\usepackage[utf8]{inputenc}

\usepackage{microtype}

\usepackage{inconsolata}
\usepackage{graphicx}
\usepackage[most]{tcolorbox}
\usepackage{enumitem}
\usepackage{listings}
\usepackage{amssymb}
\usepackage{xcolor}
\usepackage{amsmath}
\usepackage{booktabs}
\usepackage{multirow}
\usepackage{tabu}
\usepackage{arydshln}
\usepackage{array}
\usepackage{xspace,mfirstuc,tabulary}
\usepackage{comment}

\usepackage[capitalize,noabbrev,nosort]{cleveref}

\makeatletter
\crefname{section}{\S\@gobble}{\S\@gobble}
\crefname{subsection}{\S\@gobble}{\S\@gobble}
\crefname{subsubsection}{\S\@gobble}{\S\@gobble}
\crefname{figure}{Fig.}{Figs.}
\crefname{table}{Table}{Tables}

\def\paragraph{\@startsection{paragraph}{4}{\z@}{0.2em}{-1em}{\normalsize\bf}}
\makeatother

\lstdefinestyle{prompt}{
  basicstyle=\ttfamily\small,
  columns=fullflexible,
  breaklines=true,
  frame=single,
  rulecolor=\color{black!20},
  frameround=tttt,
  numbers=left,
  numberstyle=\tiny\color{black!50},
  xleftmargin=2em,
  framexleftmargin=1.5em,
  aboveskip=1em,
  belowskip=1em
}

\newcommand{\atypic}{\texttt{EventTyp}\xspace}
\newcommand{\rwtypic}{TypicFollowUp\xspace}
\newcommand{\rwatypic}{AtypicFollowUp\xspace}
\newcommand{\atypicnorm}{\texttt{EventTypGold}\xspace}
\newcommand{\nullcond}{NONE\xspace}
\newcommand{\infscore}{InfScore\xspace}

\title{Hey, guess what: CUES lead to inferences about typicality}

\author{Alison Chi$^*$ \and Sanghyeok Choi \and Hannah Rohde \and Emily Allaway \\
        University of Edinburgh \\
        $^*$Corresponding author: \texttt{a.h.chi@sms.ed.ac.uk}}

\begin{document}
\maketitle
\begin{abstract}
Throughout life, we accumulate knowledge of how typical certain events are. We use this knowledge in daily conversation, often interpreting meaning beyond what has been said explicitly.
In this work, we take steps to better understand this meaning recovery, investigating both atypicality and typicality inferences.
First, we construct two corpora, \atypic and the human-validated \atypicnorm.
We conduct large-scale human and LLM experiments and find that for humans, three \textbf{CUES} predict inferences about typicality: \textbf{C}ommonsense Knowledge, \textbf{U}tterance \textbf{E}mphasis, and \textbf{S}nark.
However, in most experimental settings, we find that LLMs lack the prerequisite \textbf{C}ommonsense Knowledge necessary to accurately assess an event's general typicality, and therefore struggle to follow these CUES.
\end{abstract}

\input{1_intro}
\input{2_related_work}

\input{3_data}
\input{4_human_llm_setup}
\input{5_cknowledge_exp}

\input{6_evenfinf_exp}
\input{7_discussion}

\section{Conclusion}
In this paper, we investigated event-related typicality inferences on a much larger scale than any previous work has before. We created \atypic and \atypicnorm, recounting the corpora creation process in detail to enable reproducibility.
We conducted a human study that gave us three CUES that predict inferences: \textbf{C}ommonsense Knowledge, \textbf{U}tterance \textbf{E}mphasis, and \textbf{S}nark. 
We asked whether LLMs are also sensitive to these CUES and found that although they sometimes are, having the Commonsense Knowledge to predict an event's general typicality is a much bigger hurdle than expected.

Our findings leave many open questions for future work. 
 For example, other features aside from textual emphasis and snark, like verb aspect or prosodic cues,
 might predict these inferences in humans. 
And for LLMs, further exploration is necessary to answer why models fail to consistently exhibit commonsense knowledge and to find solutions to mitigate this.

\section*{Limitations}
In our experiments, we tested only one form of utterance emphasis (the HeyGW setting) in eliciting pragmatic inferences. Similarly, we used a single, minimal form of dialogue context for our experiments. It remains to be seen whether other interventions, including more minimal ones tested in prior works (e.g., adding `!';~\citealp{kravtchenko2022informationally}), and other forms of context, also trigger atypicality and typicality inferences at scale for humans or LLMs. 
Furthermore, because we use textual data, our experiments cannot account for the role that prosody may play in triggering pragmatic inferences. 

The events tested in our experiments are constrained to those can that happen relatively often in an average (US) person's life, as well as relatively quickly (i.e., the event is complete within a day of starting). It remains to be seen whether other types of events can trigger typicality-related inferences in the same patterns.

For our human experiments, we specifically tested people based in the US, due to the US-centric nature of the original data. As such, the underlying participant pool may not reflect the full diversity of English speakers. Additionally, our criteria for Context and Follow-up events were developed by English speakers. Event-typicality inferences may be triggered differently for speakers of languages other than English.

In our LLM experiments, we only tested five LLM families. Since we saw varied behavior across these models, we cannot predict how LLMs from other model families (e.g., \texttt{Mistral}~\cite{jiang2023mistral7b}, \texttt{DeepSeek}~\citep{Guo_2025}) will behave. In particular, we only tested one closed source model (\texttt{GPT-5.4}). Testing with other such models could provide more insights.


\bibliography{custom}

\clearpage
\appendix

\input{appendix}

\end{document}

%% file: 1_intro.tex
\section{Introduction}
In our daily lives, we make decisions based on what we think will and will not happen. 
This knowledge about event typicality (i.e., which events are likely or unlikely to happen) helps us navigate the intricacies of human conversation, interpreting meaning beyond what speakers have literally said \cite{grice1975logic, bergen2016pragmatic}.

In this work, we investigate two types of \textbf{event-typicality inferences}. These rely on our commonsense knowledge about which events are typical (and \textit{unlikely} to be mentioned when they occur) and which are atypical (and \textit{likely} to be mentioned when they occur).
Consider \cref{fig:example}. 
We might expect that when someone thanks their boss, they typically mean it (there is no need to explicitly state this).
But we also expect that when a speaker says something, it is informative. So, when a typical event \textit{is} explicitly mentioned (left panel), our expectations of the speaker conflict with our commonsense knowledge (i.e., the speaker appears to be stating something uninformative).
To resolve this conflict, we make an \textbf{atypicality inference} that, in this context, this event that is generally typical for others is in fact \textit{atypical} for Anahita. 
Conversely, when an atypical event is explicitly mentioned (right panel), this is straightforwardly informative and so we may simply make a \textbf{typicality inference} that, in this context, this generally atypical event is \textit{more typical} for Anahita than we previously believed.\footnote{Although we call both of these phenomena \textit{inferences}, we only claim that atypicality inferences require pragmatic reasoning. We consider a typicality inference to be a belief update according to semantics (see \cref{subsec:appendix-typinf}).}

\input{figs/intro_fig}

Previous studies
have carried out only limited human experiments on event-typicality inferences 
\cite{kravtchenko2015semantically, kravtchenko2022informationally, ryzhova2023inferences}, and
even less work has been done on LLMs~\cite{kurch-etal-2024-large}.
However, event-typicality inferences pose important challenges because they rely on both commonsense knowledge and reasoning ability. 
For LLMs, investigating such inferences is particularly critical since
recent works have shown that LLMs often exhibit shallow and inconsistent commonsense event-knowledge \cite{chen-etal-2023-say, bian-etal-2024-chatgpt, kim-hwang-2025-counterfactual}. Therefore, in this paper we aim to expand our understanding of event-typicality inferences in both humans and LLMs. We ask:
\begin{itemize}[leftmargin=*,noitemsep,nolistsep]
    \item RQ1: Do humans and LLMs \textit{share Commonsense Knowledge} about event typicality?
    \item RQ2: When do \textit{humans} make inferences about event typicality?
    \item RQ3: Do \textit{LLMs} make inferences about event typicality in the contexts where humans do?
\end{itemize}

To address these, we create \textbf{\atypic}, a corpus of 2,354 labeled event tuples, and \textbf{\atypicnorm}, a subset of 858 tuples with additional human validation.
Using our corpora, we conduct large-scale experiments with both humans (${\sim}900$ participants) and LLMs (10 models and 3 inference methods).

First, our results show that LLMs mostly \textit{do not} agree with humans in their commonsense knowledge about event typicality (RQ1). This has broad implications for LLM usage, since many downstream tasks, such as task planning \cite{zhao2023large} and knowledge graph completion \cite{ji2024klr}, rely on LLMs' commonsense knowledge.

Second, when examining event-typicality inferences in humans (RQ2), we find that 
three \textbf{CUES} invite them: 1) \textbf{C}ommonsense Knowledge, 2) \textbf{U}tterance \textbf{E}mphasis (e.g., ``And hey, guess what''), and 3) \textbf{S}nark (i.e., when an atypicality inference helps the listener learn something negative about someone).

Finally, when examining LLM behaviour (RQ3) we find that while some LLMs make event-typicality inferences following similar CUES as humans, this is limited primarily to large models with Chain-of-Thought prompting~\citep{wei2022chain}.

Our contributions are: \textbf{(i)} We create two new large-scale, high-quality corpora for studying event typicality, \atypic and \atypicnorm; \textbf{(ii)} We conduct large-scale human experiments and find evidence for three CUES that can lead to inferences about event typicality; and \textbf{(iii)} We conduct extensive experiments with LLMs and show that LLMs often lack event commonsense knowledge, which undermines their ability to make event-typicality inferences.\footnote{\url{https://github.com/alisonhc/heygw-eventtyp}}

%% file: figs/intro_fig.tex
\begin{figure}[t]
\centering
\includegraphics[width=\columnwidth]{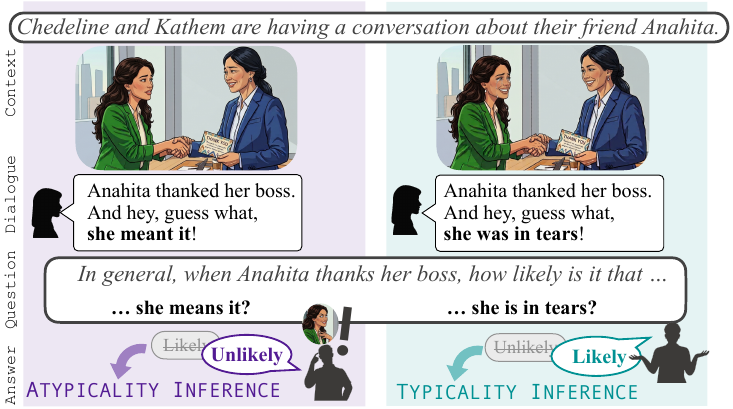}
\caption{Examples of what can cause an atypicality or typicality inference. Someone making an \textbf{atypicality} inference would answer the question on the left with \textit{unlikely}, whereas if they made a \textbf{typicality} inference, they would answer the question on the right with \textit{likely}.}
\vspace*{-0.7em}
\label{fig:example}
\end{figure}

%% file: 2_related_work.tex
 \section{Related Work}
 \paragraph{LLMs \& commonsense knowledge about events} 
LLMs have been shown to struggle to reason about relationships between events, whether they are rare/newsworthy \cite{koupaee-etal-2025-causal, islakoglu-kalo-2025-chronosense} or daily life \cite{sabour-etal-2024-emobench, Wang_Liu_Tang_Chen_2026, toroghi2026commonwhydatasetevaluatingentitybased}.
Prior work has examined LLMs' ability to differentiate event arguments~\citep{pedinotti-etal-2021-cat}, as well as how probable events are on different levels of granularity: e.g. plausible vs.~implausible \cite{jain-etal-2023-language-models, joshi2024cold} or a spectrum from probable to inconceivable \cite{kauf2023event, HU2025104640, lepori2026just}.
To our knowledge, no work has investigated the more fine-grained task of distinguishing the typicality of two events that are \textit{both highly plausible} but \textit{still different in typicality}.

\paragraph{Human pragmatic reasoning about typicality}
Pragmatic reasoning can be characterized in terms of comprehenders' expectations for what cooperative speakers are likely to do, including the expectation for speakers to provide an appropriate Quantity of information \cite{grice1975logic}. Apparent violations of this expectation (saying something ambiguous or otherwise underinformative) can invite inferences.

With atypicality inferences, a speaker's apparent uninformativity (the mention of some typical event given the context) contrasts with the listener's belief in the speaker's cooperativity. But the speaker \textit{is} informative if the reported event is actually \emph{atypical} given the context (see \cref{sec:appendix-related-work} for more details).
Existing experiments on such inferences \cite{kravtchenko2015semantically, kravtchenko2022informationally} have asked participants to read dialogue snippets and rate the mentioned events' typicality, revealing atypicality inferences for typical events but no inferences for atypical events. Subsequent work confirms the inferences reflect Quantity-driven reasoning \cite{ryzhova2023inferences}.

\paragraph{LLMs \& pragmatic reasoning}
The limited work on LLMs' pragmatic reasoning has shown that a wide gap persists between humans and LLMs \cite{hu-etal-2023-fine, 
yerukola-etal-2024-pope, 
ma-etal-2025-pragmatics}. For event-typicality inferences in particular, \citet{kurch-etal-2024-large} conducted prompting experiments with two LLM families on a set of 24 utterance tuples, which is relatively limited in scale compared to our experiments, and observed no atypicality inferences.

%% file: 3_data.tex
\input{tables/atypic-event-components}

\section{The \atypic Corpus}
\label{sec:atypic-corpus}
This section describes the process of creating the \atypic corpus.
The corpus comprises \textbf{event tuples} \textbf{(Context, \rwtypic, \rwatypic)}: Context is an event extracted from a commonsense dataset (\cref{subsec:context}), \rwtypic is a typical follow-up event to Context and \rwatypic is an atypical follow-up event to Context.
See examples in \cref{table:tuple-def}.
Both the \rwtypic{s} and \rwatypic{s} are generated by an LLM (\texttt{GPT-4o},~\citealp{openai2024gpt4ocard}). 
Both the \rwtypic and the \rwatypic depend on the Context, but we generate each \rwatypic according to both its Context and the \rwtypic. 
Therefore, after generating the \rwtypic{s} (\cref{subsec:rwtypic-gen}), we filter them to remove those that could not plausibly lead to an atypicality inference (\cref{subsec:atypicality-criteria}) and then only generate \rwatypic{s} for the remaining examples (\cref{subsec:rwatypic-gen}).
\textbf{Our final \atypic corpus has 2,354 event tuples.}

\subsection{Context Event Extraction and Filtering}
\label{subsec:context}
We source our Contexts from \texttt{ATOMIC} \cite{sap2019atomic, hwang2021comet}, a commonsense 
knowledge base
consisting of if-then event pairs (e.g., ``PersonX takes PersonY’s advice'' isAfter
``PersonY gives advice to PersonX'').
We extract all event pairs for two relation types (isBefore and isAfter) 
and limit utterances to those with up to two entities.
Combining and deduplicating them yields 33,924 utterances.\footnote{See \cref{subsec:appendix-context-extract} for why we used \texttt{ATOMIC} for Context extraction but not for follow-ups.} 
We then apply rule-based and LLM-based filtering to remove noisy or unsuitable events (e.g., idioms, passive voice; see full details in \cref{subsec:appendix-context-filter}). This leaves 12,978 Context events.

\subsection{Typical Follow-ups (\rwtypic{s}) }
\label{subsec:rwtypic-gen}

For each Context, we use an LLM 
to generate up to
three \rwtypic{s} in order to increase the diversity of our corpus (see \cref{subsec:appendix-rwtypic-gen} for details).
This yields 19,827 
(Context, \rwtypic) candidates.

\subsection{Atypicality Inference Event Criteria}\label{subsec:atypicality-criteria}
Despite prior work on atypicality inferences, no precise attributes have been identified as triggering them.\footnote{\citet{kravtchenko2015semantically} combined events from \citet{raisig2009insights}'s `scripts' with some of their own events to construct their corpus, using heuristics like `semantically uninformative' to characterize \rwtypic{s}. But these heuristics are too vague to apply in large-scale annotation.}
To create \atypic, we develop a set of specific triggering criteria through manual annotation.
We then apply these criteria through LLM-prompting to our candidate (Context, \rwtypic) pairs.

\paragraph{Criteria Development.} 
One of the authors manually annotated a subset (N = 9,188) of candidate event 
pairs
for whether or not 
it was possible to make an atypicality inference from them, and if not, why it was not possible. 
The annotations
yielded a set of 2,026 
(Context, \rwtypic) pairs
labeled as triggers for atypicality inferences. From these we determine two primary rules 
(see Appendix \cref{table:llm-filter-prompt-classes} for rule details and examples):
\begin{enumerate}[leftmargin=*, noitemsep,nolistsep] 
    \item The 
    follow-ups
    should be natural for dialogue (e.g. grammatical; `Mary takes Linda's hand' is but `Mary take Linda hand' is not).
    \item The \rwtypic should have `natural' typicality, e.g. very typical given the Context but still \emph{not entailed} by the Context. So, (``Mary gives Linda advice'', ``Linda listens'') is acceptable, but (``Mary arrives at Linda's party'', ``Linda has a party'') is not.
\end{enumerate} 

\paragraph{LLM-based Criteria Filtering.} 
We operationalize our criteria by prompting an LLM with 27 true-false questions
(see \cref{subsec:appendix-rwtypic-gen} for details)
We reject candidate event pairs where even one criteria rule is violated.
Due to inconsistency in the LLM responses, we split some 
criteria
into multiple overlapping questions in order to increase accuracy.
We use few-shot prompting with manually crafted examples and iteratively develop the prompts, favoring precision over recall. That is, we prefer the LLM rejects more good follow-ups if that ensures fewer bad ones.

To evaluate the quality of the LLM-filtering, we create 
a set of human labels by having one human annotator label 39 randomly sampled 
(Context, \rwtypic) pairs
from our dataset using the same 27-question setup as the LLM.
We then calculate precision (78.11), recall (69.93), and inter-annotator agreement in the form of Cohen's Kappa ($\kappa = .4532$)~\cite{cohen1960coefficient} between the LLM and human annotator on this set. 
While agreement is modest,
this is primarily because the LLM rejects more follow-ups than the human annotator. 
On the 
annotated sample,
the human annotator rejects 16.48\%, while the LLM annotator rejects 26.37\%.  

We run the LLM-filtering 
on all 19,827 candidate (Context, \rwtypic) event
pairs, resulting in 
4,155 that meet our criteria.

\subsection{Atypical Follow-ups (\rwatypic{s})}
\label{subsec:rwatypic-gen}
For each (Context, \rwtypic) pair 
that passes our filtering in \S\ref{subsec:atypicality-criteria},
we use an LLM to generate a candidate \rwatypic.
We instruct the model to generate a less typical, more specific alternative to the original \rwtypic (\cref{subsec:appendix-rwatypic-gen}). 
The number of generated \rwatypic{s} (4,155) is much lower than the number of \rwtypic{s} generated in \cref{subsec:rwtypic-gen} (19,827), so 
we manually validate and correct all candidate \rwatypic{s}.\footnote{Some tuples which were found to have errors were removed during this stage or retroactively, see \cref{subsec:appendix-rwatypic-gen}.}
We are left with 4,113 (Context, \rwtypic, \rwatypic) tuples, from which we subsample 2,330 so our data has only one \rwtypic and one \rwatypic per Context (note that there are up to three followups per Context from \S\ref{subsec:rwtypic-gen}, and see \cref{appendix:subsec:data-sample} for details).

We add the 24 tuples from \citet{kravtchenko2015semantically} (K\&D) to our data, rewritten to have a consistent format with our
tuples. 
In total, our \textbf{\atypic corpus has 2,354} (Context, \rwtypic, \rwatypic) event tuples. See \cref{table:tuple-def} for 
the distribution of follow-ups 
written and validated by a human or an LLM.

%% file: tables/atypic-event-components.tex
\begin{table*}
  \begin{center}
    \small
    \resizebox{\textwidth}{!}{%
    \begin{tabular}{l|l|l|rr|rr}
      \toprule 
     & \textbf{Definition} & \textbf{Example} & \multicolumn{2}{c|}{\textbf{Written by}} & \multicolumn{2}{c}{\textbf{Validated by}}\\
     & & & LLM & Human & LLM & Human\\
    \hline
    \textbf{Context} & The first event utterance & Anahita thanks her boss & \\
    \hline
      \textbf{\rwtypic} & Event utterance stating something that & Anahita means it & 2,062 & 292 & 1,226 & 1,128 \\
      & is generally \textbf{typical} given Context.  & & & &  \\
      \hline
      \textbf{\rwatypic} & Event utterance stating something that & Anahita is in tears & 1,679 & 675 & 0 & 2,354 \\
       & is generally \textbf{atypical} given Context. & & & & \\
      \bottomrule
    \end{tabular}
    }
  \end{center}
\vspace*{-0.5em}
      \caption{Example (Context, \rwtypic, \rwatypic) tuple from \atypic, along with the number of items per-type written and validated by an LLM and a human. Note that we also count the human-written and human-validated data from \citet{kravtchenko2015semantically}.
      The total number of tuples in \atypic is \textbf{2,354}.
      }
    \label{table:tuple-def}
\vspace*{-1.0em}
\end{table*}

%% file: 4_human_llm_setup.tex
\section{Human and LLM Experiment Details}\label{sec:experimentdetails}
 Here we describe our collection of human and LLM event-likelihood ratings, which applies to both our Commonsense Knowledge experiment (\cref{sec:norming}) and our Event-Typicality Inference experiment (\cref{sec:event-inference-exp}). We collect event-typicality ratings by asking how `likely' certain follow-up events are; hence, we use the term \textbf{likelihood ratings} throughout.

\subsection{Human Participant Details}
We recruit 330 participants for the Commonsense Knowledge experiment (\cref{sec:norming}) and 575 for the Event-Typicality Inference experiment (\cref{sec:event-inference-exp});\footnote{We use Prolific (\url{https://www.prolific.com/}) to recruit participants. Since \atypic's source data was created in the United States and contains US-specific references (e.g., Home Depot, Thanksgiving), we only recruit participants based in the United States.}
see \cref{appendix:human-details} for more details. 
For both experiments, each participant rated the likelihood of a batch (Context, Follow-up) pairs, where Follow-up was either \rwtypic or \rwatypic, and no Context event appeared more than once in each batch. 
Each batch contained 6 attention check questions that were randomly mixed with the other items. For each event pair, we collected likelihood ratings from multiple participants: 3 for the Commonsense Knowledge experiment and 5 for the Event-Typicality Inference experiment. 

\subsection{LLM Details}
We test 28 LLM configurations covering
three inference methods (surprisal, two prompting approaches) and five model families with two sizes each:
GPT-5.4,~\cite{openai2026gpt54systemcard}; Llama-3.1,~\cite{grattafiori2024llama}; Olmo-3,~\cite{olmo2025olmo}; Qwen3,~\citealp{yang2025qwen3}; Gemma-3,~\cite{gemmateam2025gemma3technicalreport}. See \cref{sec:appendix-llm-details} for details.
 
\input{figs/surprisal}
\paragraph{Surprisal} 
We compute surprisal (negative-log-likelihood), which can be used to probe
LLMs for underlying language understanding.
We adapt the method from \citet{cilleruelo-etal-2025-generics}, computing
surprisal $\mathrm{Sp}(\text{Completion})$ over tokens $t_i$:
$$
\mathrm{Sp}(\text{Completion})
= \sum_{i=1}^{N} -\log p\!\left(t_i \mid t_1,\ldots,t_{i-1}\right)
$$
To simulate multiple-choice question-answering, we insert each likelihood option into our input (shown in \cref{fig:surprisal-example}) and then compute surprisal on the Completion (Follow-up). Since we are only comparing surprisals of the same Follow-up, there is no need to normalize the surprisal values by length. We consider the LLM's chosen answer to be the Input likelihood option where the Completion's summed surprisal is lowest.
For this approach, we use only non-instruction-tuned (base) LLMs.

\paragraph{Prompting}
To evaluate LLMs in a way that more closely mirrors how we obtain human results, we also prompt them for a likelihood rating in the same way as we do humans.
We try both direct prompting (Prompting) and Chain-of-Thought prompting (CoT-Prompting).\footnote{We call this CoT for all LLMs, but for \texttt{GPT-5.4}, we instead set the reasoning parameter to `medium.'}
For
this approach, we use the instruction-tuned versions of the models used to compute Surprisal.

%% file: figs/surprisal.tex
\begin{figure}
    \begin{center}
    \small
    \resizebox{\columnwidth}{!}{
    \includegraphics[width=\columnwidth]{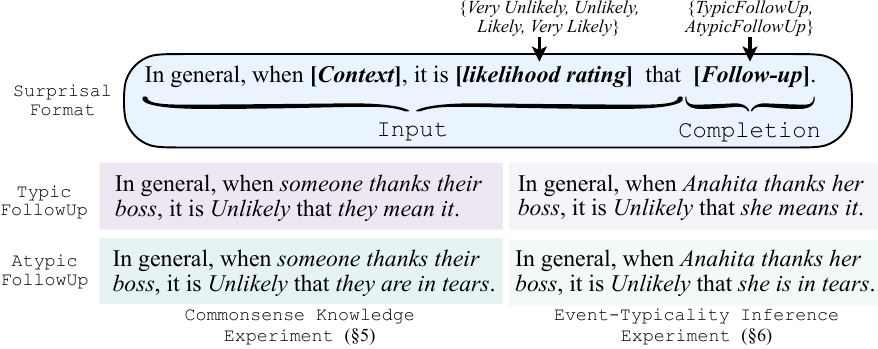}
    }
  \end{center}
  \vspace*{-1em}
  \caption{Setup for our surprisal experiments with examples using the \textit{Unlikely} likelihood rating option.
  Surprisal is computed over the Completion, given the Input. 
      }
    \label{fig:surprisal-example}
    \vspace*{-1.0em}
\end{figure}

%% file: 5_cknowledge_exp.tex
\input{tables/norming-examples}

\section{Commonsense Knowledge Experiment}
\label{sec:norming}
To answer \textbf{RQ1} (whether humans and LLMs share Commonsense Knowledge about event typicality), we elicit likelihood ratings from both humans and LLMs for every (Context, Follow-up) pair in \atypic and then compare their responses. 

We then create \textbf{\atypicnorm} (N = 858) from only the \atypic tuples where the humans voted that the \rwtypic was typical and the \rwatypic was atypical (\Cref{subsec:eventtypcold}). We use \atypicnorm for our Event-Typicality Inference experiment (\cref{sec:event-inference-exp}).

\input{tables/experiment-setup}
\subsection{Experimental Setup}
\label{subsec:normsetup}
We collect likelihood ratings for all (Context, Follow-up) pairs in \atypic following the procedures in \cref{sec:experimentdetails}. \cref{table:norm-example} shows an example of the question format we use.
For each 
tuple in \atypic,\footnote{Excluding the 24 items from K\&D, for which we manually write the typicality questions.}
we generate two questions: one for the (Context, \rwtypic) pair and the other for (Context, \rwatypic) pair. We use \texttt{GPT-4o} (\cref{subsec:appendix-cknowledge-setup}) to generate the questions, manually validating them and correcting any errors (e.g., pronoun ambiguity).
For each question, the possible 
likelihood ratings
are \{\textit{Very Unlikely, Unlikely, Likely, Very Likely}\}. 

For each LLM configuration, we calculate inter-annotator agreement (more details in \cref{sec:appendix-metrics-details}) using Cohen’s Kappa \cite{cohen1960coefficient} for the LLM and human responses (each response is 1 majority-vote). We collapse the four likelihood
options
into 
the
binary 
categories of UNLIKELY (\textit{Very Unlikely/Unlikely}) and LIKELY (\textit{Very Likely/Likely}) and calculate $\kappa ^{\text{\rwtypic}}$ for \rwtypic{s} and $\kappa ^{\text{\rwatypic}}$ for \rwatypic{s}.

\input{figs/human-llm-kappas}
\subsection{RQ1: Do humans and LLMs share Commonsense Knowledge about event typicality?}
\label{subsec:rq1}

\textit{Mostly no: humans and LLMs generally do not agree on which events are typical versus atypical.}

\cref{fig:llm-kappas} shows our agreement results.
Out of our 28 LLM configurations, only 
\texttt{GPT-5.4} with CoT has higher than $0.4$ agreement for both Follow-ups ($\kappa ^{\text{\rwtypic}} = 0.4336$ and $\kappa ^{\text{\rwatypic}} = 0.5059$). Only \texttt{GPT-5.4} and \texttt{Llama-3.1-70B-Instruct} 
achieved $\kappa > .3$ for both follow-ups with 
both prompting settings, but the other three model families 
do
not achieve 
this even 
with CoT-Prompting. This low agreement illustrates the lack of shared commonsense knowledge between our human participants and LLMs.

\subsection{The \atypicnorm Corpus}
\label{subsec:eventtypcold}
We are interested in whether \atypic's Follow-up categories (Typic/Atypic) match our human annotators' actual likelihood ratings.
We first compute, for each Context, the percentage of tuples where the average rating of the \rwtypic was higher than that of the \rwatypic (i.e., the \rwtypic was judged more likely than the \rwatypic). This condition holds for \textbf{80.04\%} of our generated tuples and 95.83\% of the K\&D tuples. This aligns with how we prompted the LLM (\cref{subsec:rwatypic-gen}), asking it to generate a Follow-up that was less likely than the \rwtypic but not necessarily unlikely.

However, for a given Context we want not only the correct \textit{relative} relation between Follow-ups but also the stricter condition that the \rwtypic is LIKELY 
and the \rwatypic UNLIKELY. 
We compute the percentage of tuples where this occurs with at least 2/3 annotators agreeing. 
From the tuples 
from \atypic that we generated, 
this condition holds for
36.09\% (841), while it does for 70.83\% (17)
tuples from K\&D,
despite the fact that 
the tuples were manually crafted by their original authors to have pairs of likely-unlikely events. 
\textbf{The 858 tuples that meet the strict condition are our final \atypicnorm corpus.}

%% file: tables/norming-examples.tex
\begin{table}
  \begin{center}
    \small
    \resizebox{\columnwidth}{!}{
    \begin{tabular}{l|ll}
      \toprule
    \textbf{Follow-up} & \textbf{Context example} & \textbf{Question} \\
     & Anahita thanks her boss & In general, when \\
     & & someone thanks \\
     & & their boss, how likely \\
     & & is it that they \\
    \midrule \midrule
    \rwtypic & Anahita means it & ...mean it? \\
    \midrule
    \rwatypic & Anahita is in tears & ...are in tears? \\
      \bottomrule 
    \end{tabular}
    }
  \end{center}
  \vspace*{-0.5em}
  \caption{Example of the questions we ask in the Commonsense Knowledge Experiment. Two questions are asked per Context, one for the likelihood of the \rwtypic and one for that of the \rwatypic.}
  \vspace*{-0.7em}
\label{table:norm-example}
\end{table}

%% file: tables/experiment-setup.tex
\begin{table*}[t]
\centering
\resizebox{\textwidth}{!}{%
\begin{tabular}{@{}l|lll|l@{}}
\toprule

& \multicolumn{4}{c}{Chedeline and Kathem are having a conversation about their friend Anahita.} \\
& \multicolumn{4}{c}{Chedeline: ``Anahita thanked her boss.''} \\
\midrule

& \multicolumn{3}{c|}{Follow-up Setting} & \\
\cmidrule(lr){2-4}

Follow-up
& \textbf{\nullcond}
& \textbf{Plain}
& \textbf{HeyGW}
& Question \\

\midrule

\textbf{\rwtypic}
&  --
& ``She meant it.''
& ``And hey, guess what,
& In general, when Anahita thanks her boss, how likely is it that \\
& & & she meant it!'' & \textit{she means it?} (Very Unlikely, Unlikely, Likely, Very Likely) \\
\hdashline
\textbf{\rwatypic}
&  --
& ``She was in tears.''
& ``And hey, guess what,
& In general, In general, when Anahita thanks her boss, how likely is it that \\
& & & she was in tears!'' & \textit{she is in tears?} (Very Unlikely, Unlikely, Likely, Very Likely) \\

\bottomrule
\end{tabular}%
}
\vspace*{-0.5em}
\caption{Setup for Event-Typicality Inference experiments (\cref{sec:event-inference-exp}). 
A participant sees 
the top dialogue (containing a Context), and ONE Follow-up setting (containing either a \rwtypic or a \rwatypic), along with its corresponding question. This means that for each tuple, there are 6 possible experiment conditions.}
\label{table:exp-setup-materials}
\vspace*{-1em}
\end{table*}

%% file: figs/human-llm-kappas.tex
\begin{figure}[t]
\centering
\includegraphics[width=\columnwidth]{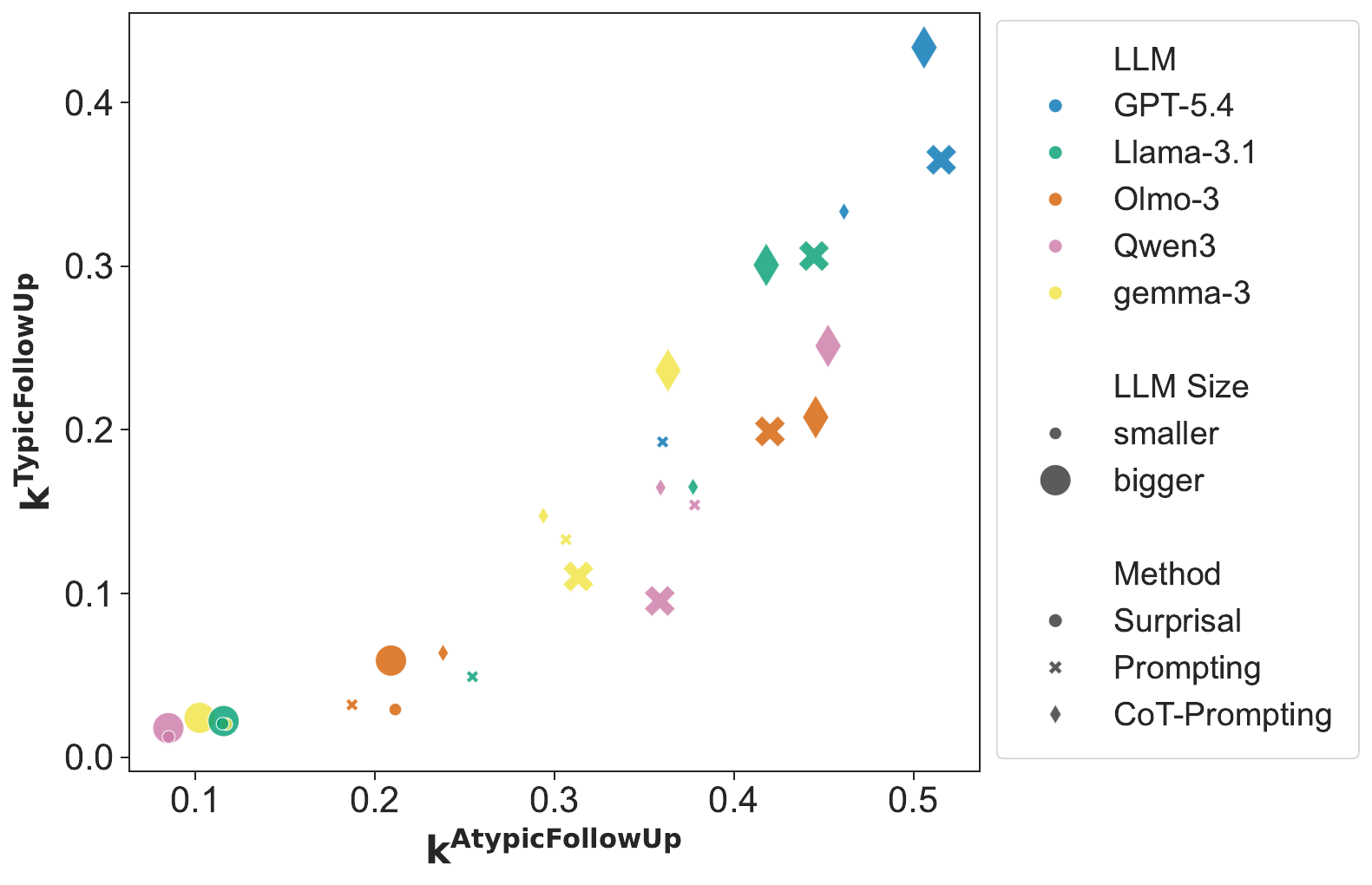}\vspace*{-0.5em}
\caption{Human-LLM agreement for the Commonsense Knowledge experiment, with $\kappa^\text{\rwatypic}$ on the X-axis and $\kappa^{\text{\rwtypic}}$ on the Y-axis.}
\label{fig:llm-kappas}
\vspace*{-0.7em}
\end{figure}

%% file: 6_evenfinf_exp.tex
\section{Event-Typicality Inference Experiment} 
\label{sec:event-inference-exp}
To answer \textbf{RQ2} and \textbf{RQ3}, we collect human and LLM likelihood ratings on a sample (N = 450) of \atypicnorm with \textit{dialogue scenarios}. We subsample due to budget constraints
(see \cref{subsec:appendix-eventtyp-exp-sample}).

\subsection{Experimental Setup}
\paragraph{Dialogue Scenarios} To create an environment where pragmatic inferences are possible, we insert events into dialogue scenarios and then ask questions about those scenarios. Unlike \citet{kravtchenko2022informationally}, we opt to 
use minimal dialogue framing
in order to avoid confounding factors. 
We create two example instances from each (Context, \rwtypic, \rwatypic) tuple using randomly selected rare names (see \cref{subsec:appendix-names} for details). Then we use \texttt{GPT-4o} to generate a typicality question for the \rwtypic and \rwatypic events. We manually validate and correct all questions. Unlike in the Commonsense Knowledge experiment (\cref{sec:norming}), the questions 
are 
use names specific to the dialogue scenarios.

\paragraph{Experimental Conditions} \label{subsec:exp-design} 
We embed the Context in a dialogue scenario and then
investigate how the likelihood ratings for its Follow-up (\rwtypic/\rwatypic) change when the Follow-up is mentioned (\textbf{Plain} and \textbf{HeyGW}) or unmentioned (\textbf{\nullcond}). \cref{table:exp-setup-materials} explains this setup.
In the \textbf{Plain} setting, the Follow-up is mentioned without Utterance Emphasis, while in the \textbf{HeyGW} setting the Follow-up is preceded by Utterance Emphasis (the phrase ``And hey, guess what''). 
We consider a participant's response in the \nullcond 
setting
to be a control (i.e., the Follow-up's baseline likelihood given the Context). Our analyses compare responses from the \nullcond 
setting
to those from Plain or HeyGW.

\paragraph{\textbf{Inference score (\infscore)}}
To compute whether an atypicality or typicality inference was made, we define 
\infscore (see \cref{sec:appendix-metrics-details}). 
For each individual event, \infscore measures the difference in average responses between the \nullcond and the mentioned (Plain or HeyGW) settings.
A positive \infscore indicates an atypicality inference and a negative \infscore a typicality inference. The magnitude of \infscore represents inference strength.

\paragraph{Snark Annotation}
We hypothesize that one possible reason why speakers produce utterances that lead to \textbf{atypicality} inferences is that they wish to convey negative information about someone without it being \textit{explicitly} negative. 
We define these scenarios as \textbf{Snarky}: if the atypicality inference helps the lister learn that the person being talked about has some negative trait, and the listener can therefore interpret the speaker as having a snarky attitude.
For example, on the \rwtypic side of \cref{fig:example}, if an atypicality inference is made (i.e., that Anahita usually does not mean it when she thanks her boss), this may mean Anahita is fake or dishonest. Thus, we label this scenario as \textit{snarky}.
For all (Context, \rwtypic) pairs in \atypicnorm, we manually annotate whether they form a snarky scenario or not. See \cref{subsec:appendix-snarky-more} for a deeper discussion on how snark relates to commonsense knowledge and irony.

\input{figs/human-results}
\input{figs/human-shady}

\subsection{RQ2: When do humans make inferences about event typicality?}
\label{subsec:rq2}

\paragraph{Atypicality inferences} \textit{We find that humans only consistently make atypicality inferences on \rwtypic{s} when Utterance Emphasis is present;
Snark increases these inferences.}

\cref{fig:human-results-plot} shows the \infscore{s} for
the Plain vs. HeyGW 
settings.
We can see that for \rwtypic{s}, $\text{\infscore}> 0$ appears often for HeyGW
(p$<$0.001, see \cref{subsec:appendix-ordinal}), indicating atypicality inferences are made when Utterance Emphasis is present. 
But for Plain,
the opposite, $\text{\infscore}< 0$, is more frequent (p$<$0.001), indicating \textit{typicality} inferences.
This contrasts with \citet{kravtchenko2022informationally}'s work, which found that participants made \textit{a}typicality inferences in the Plain setting. A possible reason for this is that their events were presented within
verbose backstories explicitly mentioning typicality, which may have primed participants to make atypicality inferences.
Our items, in contrast, use only a short dialogue.

To check Snark's effect on atypicality inferences, we focus on \rwtypic events in the HeyGW
setting,
the condition for which participants made consistent atypicality inferences. \cref{fig:human-shady-plot} shows that the difference between ratings in the \nullcond and HeyGW settings is higher when 
Snark
is present (p$<$0.001, see \cref{subsec:appendix-ordinal}), confirming
that participants make more atypicality inferences for snarky 
scenarios.
This finding is consistent with the claim that people use pragmatics to communicate about socially sensitive situations.
But for a listener to recognize `snark' in a specific scenario and make an atypicality inference on it,
they must not only have the commonsense knowledge that the \rwtypic is generally typical, but also that it \textit{should} happen according to social conventions or expectations of competence (see \cref{subsec:appendix-snarky-more}). So, it is interesting that these scenarios are the ones with higher rates of atypicality inferences.

\paragraph{Typicality inferences} \textit{We find that humans make typicality inferences for \rwatypic{s} regardless of whether Utterance Emphasis is present;
humans
also make them for \rwtypic{s} when Utterance Emphasis is not present.}

For the majority of \rwatypic{s}, $\text{\infscore}< 0$ (\cref{fig:human-results-plot}), showing typicality inferences for both the Plain and HeyGW settings (p$<$0.001).
This not consistent with \citet{kravtchenko2022informationally}'s findings, which 
showed
no significant difference between likelihood ratings for 
unmentioned vs. mentioned Follow-up
settings.
However, our finding \textit{is} consistent with the intuition that if a Follow-up event is initially unexpected (i.e., has a low likelihood rating in the \nullcond setting), but a listener learns that it happened once (i.e., mentioned in Plain/HeyGW settings), the listener may assume that it actually happens more often than they initially expected
. In other words, they simply update their beliefs based on what happened without making a pragmatic inference.

\input{figs/llm-heygw-typ-atyp-percent}

\subsection{RQ3: Do LLMs make inferences about event typicality in the contexts where humans do?}
\label{subsec:rq3}
\textit{Only large models with CoT-Prompting do this.}

We focus our analyses on the HeyGW setting, as this is 
where humans made both typicality and atypicality inferences in the expected event contexts (see \cref{subsec:appendix-ordinal}).
\cref{fig:llm-heygw-typ-atyp-percent} shows that very few LLM configurations achieve a similar balance to humans between typicality and atypicality inferences. We confirm this with statistical tests \cref{subsec:appendix-ordinal}: only 6 out of the 28 configurations produce statistically significant inferences (i.e., the \textit{mentioned} status of the Follow-up in the dialogue, e.g., \nullcond vs. HeyGW, predicts the Follow-up's likelihood rating) for both inference types, and 4 out of those are with CoT prompting on larger models.
For most larger LLM configurations which showed atypicality inferences, we also see an increase in these inferences for Snarky scenarios (\cref{subsec:appendix-ordinal} and \cref{fig:llm-shady-plot}).

%% file: figs/human-results.tex
\begin{figure*}[!t]
\centering
    \centerline{\includegraphics[scale=0.6] {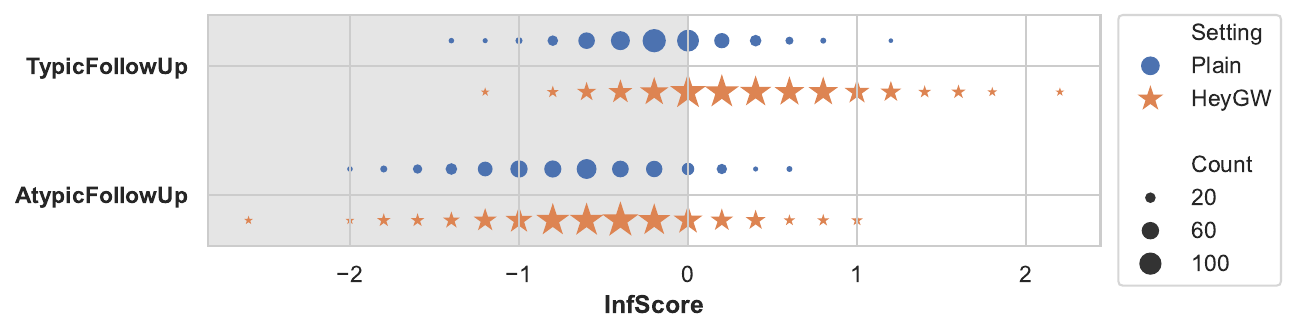}}
    \vspace*{-0.7em}
    \caption{Human results for \rwtypic{s } vs. \rwatypic{s}, where a point's size denotes the number of {\infscore}s equal to its X value.
    \infscore > 0 = \textit{atypicality} inference, and \infscore < 0 = \textit{typicality} inference.}
    \label{fig:human-results-plot}
    \vspace{-0.8em}
\end{figure*} 

%% file: figs/human-shady.tex
\begin{figure}[t]
\centering
    \centerline{\includegraphics[scale=0.3] {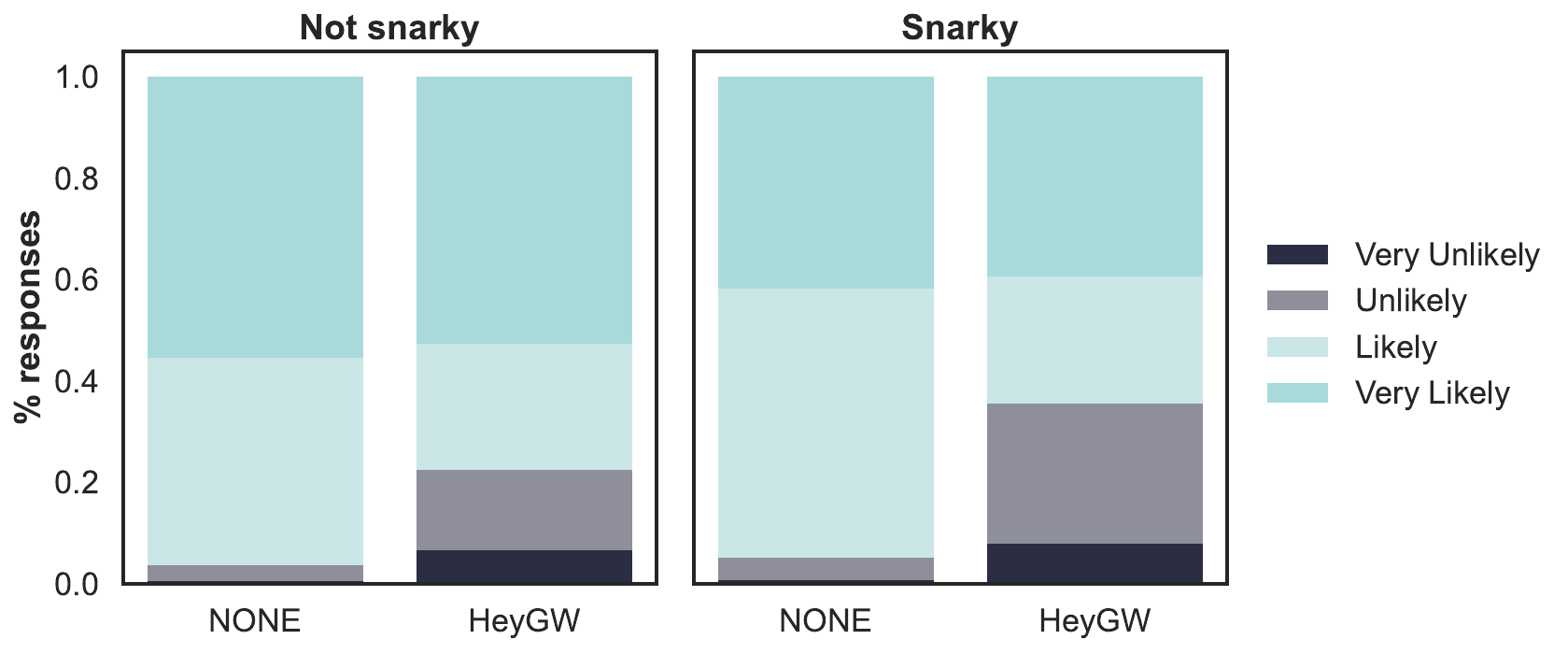}}
    \vspace*{-0.7em}
    \caption{Human distributions of responses in the \nullcond vs. HeyGW settings for \rwtypic{s}, with non-snarky
    scenarios (left)
    and snarky 
    scenarios (right).}
    \vspace*{-0.8em}
    \label{fig:human-shady-plot}
\end{figure} 

%% file: figs/llm-heygw-typ-atyp-percent.tex
\begin{figure}[t]
\centering
\includegraphics[width=\columnwidth]{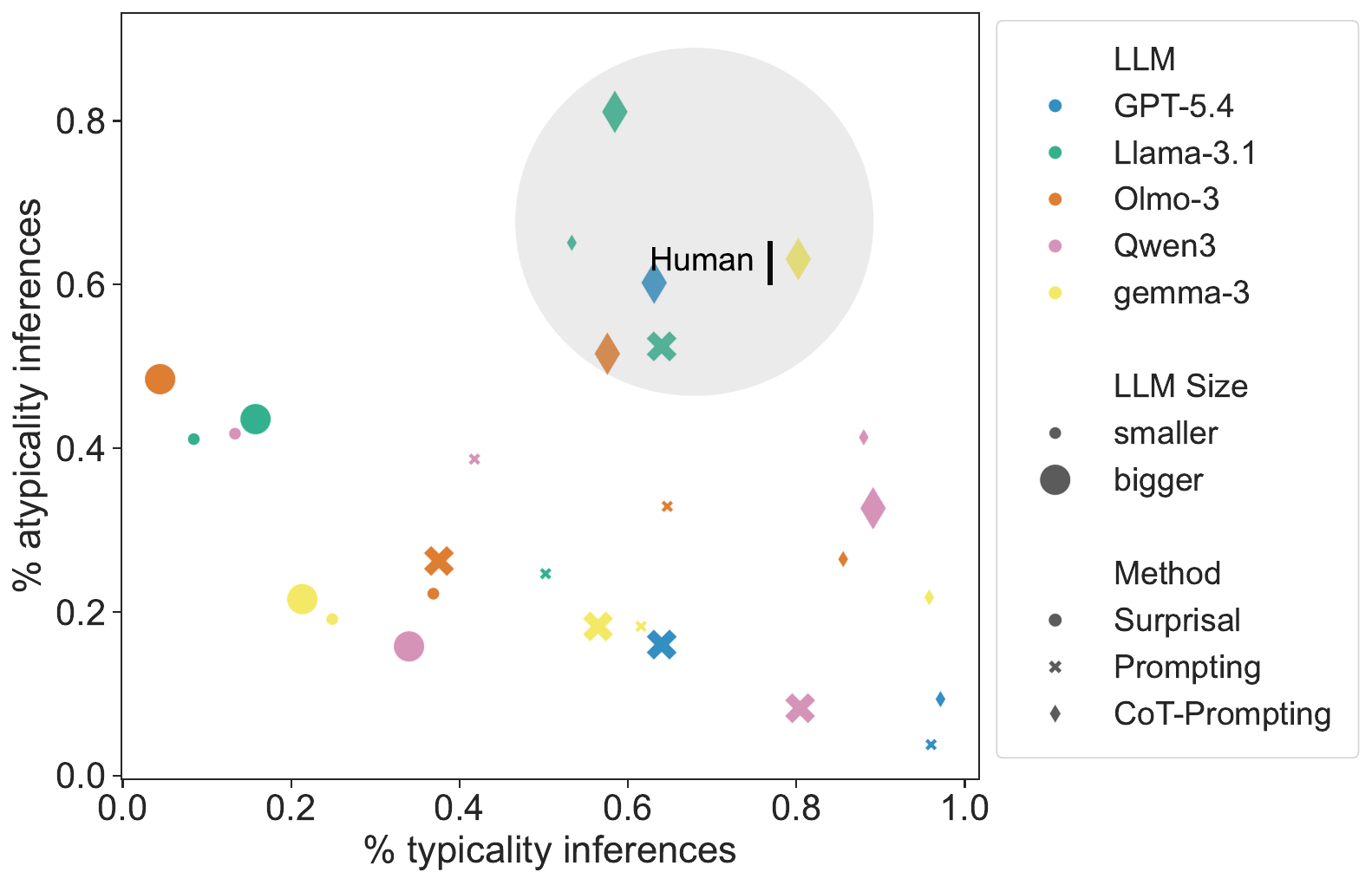}\vspace*{-0.5em}
\caption{Percentage of Follow-ups in the HeyGW setting
where an LLM made a typicality (X-axis) vs. atypicality inference (Y-axis). Points in the gray halo are LLM configurations where the atypicality and typicality inferences were both statistically significant (p$<$0.01).}
\label{fig:llm-heygw-typ-atyp-percent}
\vspace*{-0.5em}
\end{figure}

%% file: 7_discussion.tex
\input{figs/llm-norm-vs-mainexp}
\section{Discussion}
While some LLMs appear to follow the same CUES as humans in making event-typicality inferences, the story may be more complicated 
due
to their often-inconsistent displays of commonsense knowledge.
With LLMs, we find a significant (p$<$0.01) correlation between higher commonsense knowledge and more event-typicality inferences (see \cref{subsec:appendix-figs}), which indicates that it is an essential component of these inferences.
However, LLMs also fail to exhibit consistent event likelihood ratings, 
calling into question their commonsense knowledge. \cref{fig:llm-norm-vs-none} shows the percentage of
Follow-ups where
each LLM
configuration rated the likelihood of the that Follow-up \textit{the same} in both the Commonsense Knowledge Experiment (\cref{sec:norming}) and the Event-Typicality Inference Experiment (in the comparable NONE-setting; \cref{sec:event-inference-exp}).
The LLMs' consistency is very low compared to humans: no LLM configuration apart from \texttt{GPT-5.4} with CoT scored higher than 80\% consistency on both \rwtypic{s} and \rwatypic{s}.

Note that we would hope for much higher scores because in each case, the questions are about \textit{exact same} Follow-up events; the only difference is the latter's addition of dialogue and named entities.
The fact that LLMs show such inconsistency when asked about the exact same events suggests that they are more sensitive to the \textit{framing of the prompt} than to the \textit{actual typicality} of an event (see \cref{appendix:discussion}).
This is in line with prior work on both commonsense knowledge tasks and more general tasks, which has found that when the wording of the prompt is slightly changed, LLMs give inconsistent, and sometimes even opposite, answers to the same exact questions \cite{chen-etal-2023-say, bian-etal-2024-chatgpt, kim-hwang-2025-counterfactual, liu2025aligninglogicmeasuringevaluating}.

%% file: figs/llm-norm-vs-mainexp.tex
\begin{figure}[t]
\centering
\includegraphics[width=\columnwidth]{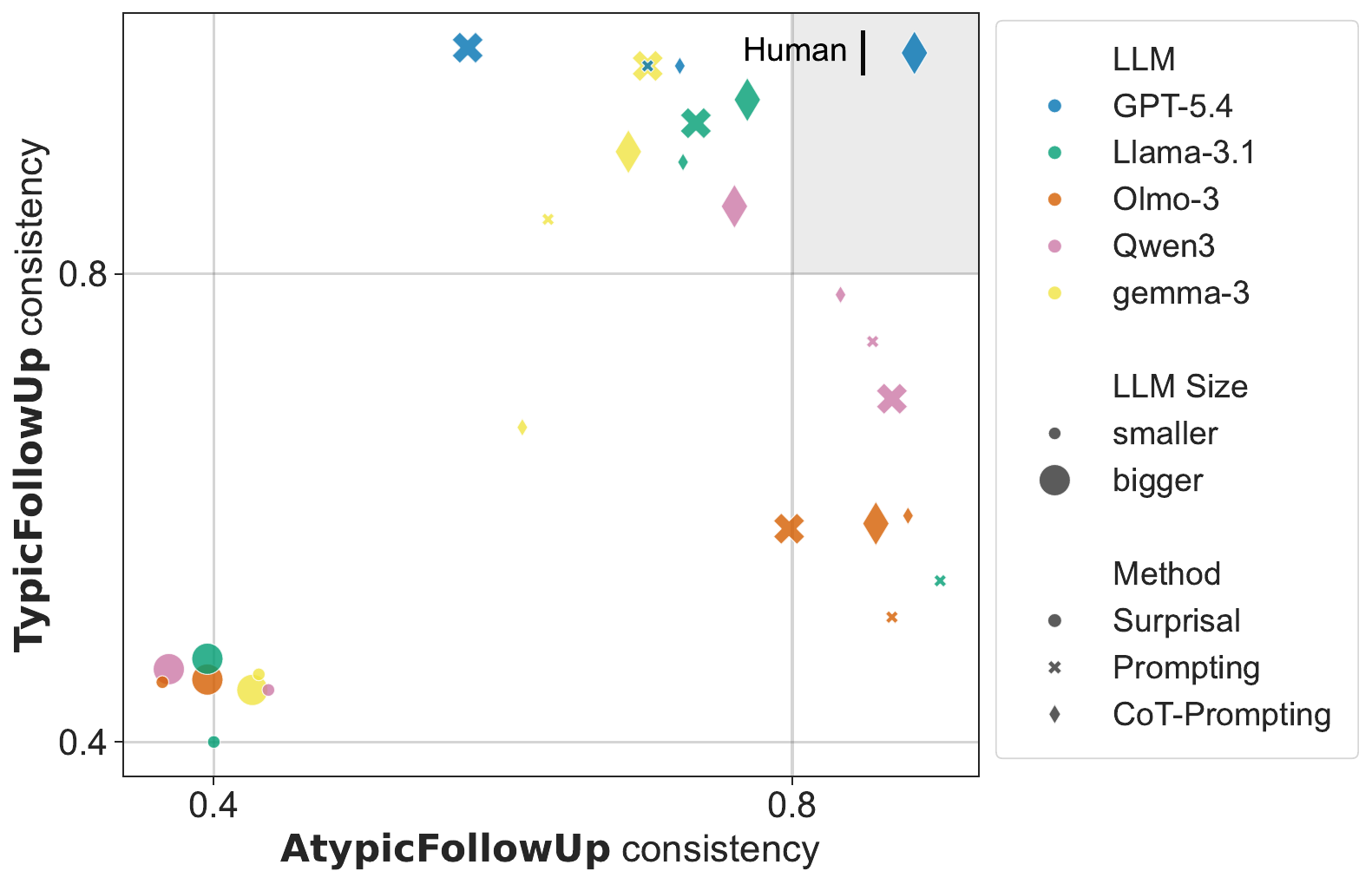}\vspace*{-0.5em}
\caption{LLM consistency; i.e. the percentage of Follow-ups where responses from the Event-Typicality Inference experiment \nullcond setting matched those from the Commonsense Knowledge experiment, (X-axis for \rwatypic{s} and Y-Axis for \rwtypic{s}).}
\label{fig:llm-norm-vs-none}
\vspace*{-0.7em}
\end{figure}

%% file: appendix.tex
\section{Pragmatics Background}
\label{sec:appendix-related-work}
\subsection{Cooperativity \& Quantity}
Pragmatics is the branch of Linguistics that targets questions about how speakers package information and how listeners draw inferences and recover intended meaning. In assessing why a speaker produces a particular utterance, listeners bring to bear various expectations about the ways in which the speaker's contribution will be relevant and, crucially for our purposes, newsworthy.  

These expectations reflect a set of principles that are posited to underlie cooperative communication \cite{grice1975logic}: A speaker's contribution is expected to be (i) appropriate linked to the discourse topic and goals (via some Relation to prior information), (ii) sufficiently informative (as measured in the Quantity of new information), and also (iii) truthful (to uphold the Quality of the contribution). Given these expectations for cooperativity, a speaker who makes an apparent violation of any of these can invite subsequent inferences, i.e., how does the discourse contribution uphold a commitment to cooperativity despite the apparent violation of Quality/Relation/Quantity? For example, an apparent violation of Quality may be interpreted as sarcastic, or an apparent violation of relation may encourage the listener to seek additional pathways to relevance. When a speaker's contribution appears to violate Quantity, the listener may potentially infer speaker ignorance (if the speaker has provided unexpectedly little information) or they may seek to establish why the contribution is newsworthy at all (if the speaker has been unexpectedly redundant).  

Quantity-driven reasoning includes the treatment of scalar implicatures, whereby the use of an ambiguous scalar (e.g., `warm') generates an implicature from its seeming underinformativity. A comprehender reasons that if the speaker is cooperative, their choice to not use a more informative alternative (`hot') implies the stronger meaning is not true ($\rightarrow$ warm but not hot).  That comprehenders seek to establish a cooperative interpretation is what drives a range of Quantity inferences about apparent redundancy \cite{degen2020redundancy, rohde2022color}, underspecification \cite{franke:2011:quantity}, or triviality \cite{rees2026you}, including atypicality inferences.

\subsection{Atypicality Inferences}
With atypicality inferences, a speaker's apparent uninformativity (the mention of some individual's \rwtypic act), combines with the listener's belief that the speaker is cooperative: The speaker \textit{is} informative if the reported act is actually \emph{atypical} for that individual.  

For the example in Figure \ref{fig:example}, a speaker asserts content that is typically inferable from shared knowledge of day-to-day events: When thanking one's boss, it is typical for an individual to mean it. A speaker who states that information explicitly ("XX thanked their boss. [And hey, guess what,] she meant it [!]") appears to violate expectations about the appropriate Quantity of information that should be present in a cooperative contribution to a discourse.  If the listener trusts that the speaker intends to be cooperative, they may look for ways in which the contribution is indeed informative. A such, the listener may `rescue' the speaker's seemingly redundant contribution by drawing an atypicality inference:  For this individual, the mentioned event \textbf{is} newsworthy and the speaker intends to convey that the individual's behavior is atypical \textbf{for that individual} (whose usual behavior is at odds with accepted norms for such situations).

Atypicality inferences are thus an example of non-trivial pragmatic reasoning.  They require the listener to draw a conclusion (that the individual doesn't typically do X) from a statement whose literal meaning provides evidence of the opposite (that the individual *has* done X). In this way, they depend on subtle reasoning that reflects broader expectations of speaker cooperativity alongside an awareness that what a speaker intends to convey may go beyond what has been explicitly said.

\subsection{Typicality inferences}
\label{subsec:appendix-typinf}
Unlike in atypicality inferences, typicality inferences do not appear to be pragmatic in nature. There is no apparent redundancy when a listener hears "Anahita thanked her boss. And hey, guess what, she was in tears [!]" Being in tears while thanking one's boss is generally atypical and therefore aligns with listeners' expectations to contribute the appropriate Quantity of information to the discourse. 
Upon learning this new information about Anahita, the listener may update their beliefs and think that for Anahita, crying is actually more typical than it is for others. 
As such, we do not consider typicality inferences to be pragmatics; but rather a straightforward belief update.

\section{The \atypic Corpus} \label{sec:appendix-data}
Note that for all LLM generation and filtering steps, we use \texttt{GPT-4o}, specifically `gpt-4o-2024-08-06,' with a temperature of 1. We selected this model specifically because it was the best-performing model that was within our budget.

\subsection{Context Event Extraction} \label{subsec:appendix-context-extract}
\paragraph{Source data.} We source our context events from \texttt{ATOMIC} \cite{sap2019atomic, hwang2021comet}, a commonsense reasoning dataset consisting of if-then event pairs.
We limit events to those with up to two named Person entities (PersonX and PersonY). 

\paragraph{We cannot directly use event relations from \texttt{ATOMIC}.}
We originally wanted to see if we could directly pull (Context, \rwtypic) and (Context, \rwatypic) pairs from \texttt{ATOMIC}, but this does not work because:
\begin{itemize}[noitemsep, nolistsep]
    \item After manually looking at some of the events, we determine that the proportion of \rwtypic{s} that would meet our criteria (see \cref{subsec:atypicality-criteria} and \cref{subsec:appendix-rwtypic-gen}) is too low. For example, highly atypical events like ``PersonX murders PersonY'' are quite common, as well as directly entailed events like ``PersonX takes PersonY's advice'' isAfter ``PersonY gives advice to PersonX''. Although our LLM-generated follow-ups still have errors, the filtering did not need to eliminate as much.
    
    \item We wanted one \rwtypic and one \rwatypic for every Context, but that does not follow the structure of \texttt{ATOMIC}, where one Context might appear any number of times, possibly only mapping to one follow-up type (\rwtypic or \rwatypic) each of those times.

    \item Aggregating all unique events from the isBefore/isAfter relation type, regardless of position, enabled us to start with the largest possible pool of Contexts from which to generate follow-ups.
\end{itemize}
So, regardless of whether an event was originally considered the `before' or the `after' event in a pair, we put all events into the same pool of Contexts which we later use to generate the \rwtypic{s} and \rwatypic{s}. 

\input{tables/rwtypic-examples}

\subsection{Context Event Filtering} \label{subsec:appendix-context-filter}
The \texttt{ATOMIC} dataset is noisy, so we use both rules-based and LLM-based methods to filter it. We replace all instances of PersonX with Mary and PersonY with Linda, replacing all pronouns with she/her/hers. We perform LLM-filtering in two passes using \texttt{GPT-4o}: the easy pass and the hard pass. The \textbf{easy pass} consists of filtering out utterances with incomplete or otherwise malformed clauses, idioms, passive voice, implicit verbs, non-present tense, pronouns with no antecedents, qualitative statements, and statements explicitly about habituality.
The \textbf{hard pass} attempts to remove items that are too vague, make absolutely no sense without context, or describe extreme situations like murder.
The result is 12,978 Contexts.

\subsubsection{Easy pass prompt}
\input{prompts/initial_filtering/classify_easy}
\subsubsection{Hard pass prompt}
\input{prompts/initial_filtering/classify_hard}

\subsection{Typical Follow-ups (\rwtypic{s})} \label{subsec:appendix-rwtypic-gen}
We prompt \texttt{GPT-4o} to generate up to three \rwtypic{s}: a state of being, subaction, and next-action (see \cref{table:category-def} for examples). This approach increases the diversity of our dataset compared to \citeauthor{kravtchenko2015semantically}'s (\citeyear{kravtchenko2015semantically}) corpus, which consists of mainly subaction or next-action tuples.
Although some state-of-being utterances may not be considered `events,' for simplicity, we use the term `event' throughout.
We allow the model to return None at its discretion (\cref{prompt:gen3clause}). 

\input{prompts/rwtypic_gen/gen3clauses}

\paragraph{LLM-based Criteria Filtering} 

\input{tables/rejection_twoclass}
\cref{table:llm-filter-prompt-classes} describes each prompt we use and links to our specific prompts in \cref{sec:appendix-all-prompts}. Note that even if, during the prompting process, an LLM's answer warrants rejection of a follow-up, we still prompt it with all 27 prompts. This is to a) allow us to go back and change individual prompts if we want to and b) remove pairs of answers when they contradict each other.

Specifically, a logical LLM should never answer entailment (too likely, \cref{subsec:entail}) prompts and Expectedness (not likely enough) (\cref{subsec:appendix-expect}) prompts the same way. So, when this happens, we remov the LLMs' answers to both of these sets of prompts.

Similarly, we had originally tried Explanation (\cref{prompt:explain}) and Result (\cref{prompt:result-result-ab}, \cref{prompt:result-result-ba}) prompts. However, we find that the LLM is especially bad at assessing causality, and answers the three prompts extremely inconsistently. Since almost every set of answers contradicts each other, the prompts have no effect in the end.

\paragraph{Generating typicality questions used for \rwtypic filtering}
For the prompt we used to filter out (Context, \rwtypic) pairs which had no typicality (\cref{prompt:typicality-q}), we input a question about typicality to the LLM and ask it if the question makes sense (\cref{prompt:gen-often-q}). In order to do this, we first separately prompt the LLM to generate the typicality question in the form of:
``When \{Context\}, how often \{{\rwtypic}\}?''
We find that this question format forces a frequency interpretation unlike the ``In general, when...how likely'' format used in our later experiments, and this frequency interpretation makes the question sound infelicitous when the event has no typicality. So, that is why we choose this question format.
\input{prompts/llm_filtering/gen-often-q}

\subsection{Atypical Follow-ups (\rwatypic{s})} \label{subsec:appendix-rwatypic-gen}
\paragraph{Generation} Below are the prompts (\cref{prompt:gen-rwatypic-state} for states, \cref{prompt:gen-rwatypic-subaction} for subactions, and \cref{prompt:gen-rwatypic-next} for next-actions) used for \rwatypic generation. Both the state and subaction prompts originally had errors where a few of the examples were not joined properly with the newline character, and this could have degraded LLM performance due to tokenization problems. However, since all generated \rwatypic{s} were manually filtered, this should not have caused any downstream problems.

\input{prompts/rwatypic_gen/gen-rwatypic-state}
\input{prompts/rwatypic_gen/gen-rwatypic-subaction}
\input{prompts/rwatypic_gen/gen-rwatypic-next}

\paragraph{Manual validation, correction, \& removal}
We choose to manually validate our generated \rwatypic{s} in order to minimize error propagation (as the \rwatypic{s} were generated from already LLM-filtered Context-{\rwtypic} pairs). Since we have close to 4,000 generated events (as opposed to over 12,000 or \rwtypic{s}), one of the paper authors had enough time to do the validation.

During the \rwatypic validation, we find that some of the {\rwtypic}s that had passed earlier filtering had errors. So, in those cases, we also correct the \rwtypic{s} when possible.

But sometimes, it isn't possible to correct the \rwtypic, because the \textit{Context} it depends on was problematic (e.g. the filtering from \cref{subsec:appendix-context-filter} had not filtered it out). In those cases, we just remove the tuple entirely. We find some problematic cases during the later manual validation stage (see \cref{subsec:appendix-cknowledge-setup}) and remove them retroactively. In our released data, we trace each generation and filtering stage and provide a list of which tuples got retroactively removed.

\subsection{Data Sample}
\label{appendix:subsec:data-sample}
After the \rwatypic validation and filtering, we are left with 4,113 tuples. Since we had originally generated up to 3 follow-ups per Context (state/subaction/next-action), there are some duplicate Contexts. We sample so that each Context is unique, oversampling states and subactions to try to alleviate imbalances in the original data, but the resulting sample still has many more next-actions than the other two categories. We also use \texttt{GPT-4o} to extract the main `action' from each Context utterance and further deduplicate those, only keeping one Context per action. Nonetheless, this action extraction was not strict, so some Context in \atypic are still about the same topic, e.g. ``camp with friends'' and ``go camping'' might both be kept. The result is 2,330 tuples, upon which we add the 24 from \citet{kravtchenko2015semantically}.

\section{Human and LLM Details}
\subsection{Human Participant Details}
\label{appendix:human-details}
\input{tables/attention-checks}
Below are more details about our human participants. All participants were US-based, had at least a 98\% approval rating on Prolific, had at least 100 previous submissions, and stated that their first and primary language were both English. Our aim was to pay equivalent to 
\$12.50/hr,
but since we paid per task and not per hour, the estimation was not perfectly accurate, and we averaged out to paying more.

\cref{table:attention-checks} shows our exact attention check questions for both the Commonsense Knowledge and the Event-Typicality Inference experiments. Before deciding on them, we informally piloted each one in order to ensure that any participant who was paying attention could answer them correctly. We format the questions to blend in with the rest of the items in their corresponding study. In each study, all participants see the exact same attention checks.

\paragraph{Commonsense Knowledge Experiment} We recruited 354 participants in total, and we required 330 complete submissions (3 * 2 pilot batches of 24, 3 * 108 larger batches of 44.) Each batch also included 6 attention checks (see \cref{appendix:human-details}), so we had 30 questions for the pilot batches and 50 for the rest.

340 were approved, 12 returned, 1 timed out, and 1 was rejected. Out of the 340 approved, 332 had usable submissions (i.e., passed the attention checks); however, due to a an error in our processing, we overwrote data from 2 participants and had to recruit 2 replacements. For this study, we had auto-approve enabled, and therefore were only able to reject one participant who failed attention checks but was flagged by Prolific as having completed the study too quickly. This means 9 people total failed the attention checks, but 8 were approved due to their responses being completed in a normal amount of time and therefore not being flagged by the Prolific system.

\paragraph{Event-Typicality Inference Experiment}
For this experiment, we only recruited participants who had passed a pragmatics pre-test (see \cref{appendix:pretest} for details).
Due to batching, each participant answered between 30 and 32 questions, consisting of about 24 questions from \atypicnorm and 6 attention checks. All participants see the same attention checks. We randomized and divide the data so that each question is answered by 5 different participants and no participant sees any Context utterance more than once.

For our data collection to be complete, we needed responses from 540 participants who passed our attention checks. So, our policy was to recruit replacements for any participants who failed the attention checks. Out of the 676 possible participants, 634 attempted the main experiment.
Of these attempts, 575 resulted in approval,\footnote{Includes the 35 who got 4/6 of the attention checks correct, a failing score but not enough for us to reject.} 39 resulted in a return, 19 resulted in rejection, and 1 timed out. Our pay averaged 
\$21.50
over all rounds of the experiment.

\subsection{LLM Details}
 \label{sec:appendix-llm-details}
\input{tables/all_llms}
\cref{table:llms-used} describes which LLMs we choose to use, as well as their sizes and whether we used the base versions, instruct versions, or both. We do not change the model parameters.
Since our inference is performed on so little data (N = 4,708) for the Commonsenes Experiment and (N = 2,700) for the Event-Typicality Inference experiment, it takes about 5 minutes for us to run Prompting inference on NVIDIA A100 GPUs and about 30 minutes for CoT-Prompting. For Surprisal, it usually took between 5-30 minutes, as we did not optimize with vLLM and directly used the transformers library.

\subsubsection{LLM version selection} \label{subsubsec:olmo-qwen}
Our goal is to keep each LLM configuration as consistent as possible across open-weight LLMs. This means our goal is to select models where both a base and instruction-tuned version exist, and for the instruction-tuned version, we wanted to avoid using Thinking so we could use the CoT prompting approach across models. Therefore, we made two decisions for Olmo-3 and Qwen3 specifically.
\begin{itemize}[leftmargin=*, noitemsep, nolistsep]
    \item \textbf{Olmo-3} \cite{olmo2025olmo}: Olmo-3 has an \textit{Instruct} model for its smaller version, but its larger model (32B) uses Thinking which cannot be switched off. So, for the larger model with prompting, we used Olmo-{\textbf{3.1}}-Instruct, whose base version is the same as for Olmo-3. 
    \item \textbf{Qwen3} \cite{yang2025qwen3}: For our instruction-tuned versions, we chose Qwen3-30B-A3B and Qwen3-8B, but we disabled thinking for both our CoT and non-CoT prompting configurations. This is because we wanted a fair comparison with other models with our own CoT prompt (see \cref{subsubsec:prompting-details} and \cref{prompt:CoT-response} for an example).
\end{itemize}

\subsubsection{Detailed Setup of Prompting} \label{subsubsec:prompting-details}
To create our prompts, we append the text ``Choose one of the following options: `Very Unlikely', `Unlikely', `Likely', `Very Likely' to the end of each question in \atypicnorm.
We find this to be necessary in order to prevent unwanted bias toward `Likely' that is introduced by the prompt (``...how  \emph{likely} is it ...'').

We sample 20 answers per prompt with temperature 1.0 for all model families except \texttt{GPT-5.4}, for which we sample 10. We reverse the response-option order for half of the samples to minimize order bias.

We consider with or without CoT for the prompting experiment. Here, we provide details for each setting, along with an example prompt and the LLM's response.

\begin{itemize}[leftmargin=*, noitemsep, nolistsep]
    \item \textbf{Without CoT.} For open-source models, we force the LLM to directly output the answer by constraining the output space to the four answer options. For \texttt{GPT-5.4}, we set `reasoning\_effort' to none to disable reasoning.\footnote{\url{https://developers.openai.com/api/docs/guides/reasoning}}
    \item \textbf{With CoT.} For open-source models, we force the LLM to generate reasoning about its response first and then give its final answer. We achieve this by using ``structured output'' generation from vLLM \citep{kwon2023efficient}. For \texttt{GPT-5.4}, we set `reasoning\_effort' to medium.
\end{itemize}

\input{prompts/prompting/without-reasoning}

\input{prompts/prompting/with-reasoning}

\section{Commonsense Knowledge Experiment} \label{sec:appendix-norm}

\subsection{Experimental Setup} \label{subsec:appendix-cknowledge-setup}
\cref{fig:norm-ui} shows an example of a likelihood rating question in the user interface format our human participants saw.
We use the prompt in \cref{prompt:gen-norm-anon-q} to generate the likelihood questions. Note that the actual raw output is a question in the format of:

\begin{quote}
    When \{Context\}, how likely \{Follow-up\}?
\end{quote}

We then use rules-based methods to pre-pend `In general, ' to the question and lowercase the `When.' For each resulting question, each human answer validates it and corrects an errors (e.g., pronoun ambiguity).

\input{prompts/norming/gen-norm-anon-q}
\input{figs/norm_screenshot}

\input{figs/mainexp_screenshot}

\section{Event-Typicality Inference Experiment} \label{sec:appendix-eventtyp-exp}
\cref{fig:mainexp-screenshot} shows an example of a likelihood rating question from \atypicnorm in the format that our human participants saw.

\subsection{Data Sample} \label{subsec:appendix-eventtyp-exp-sample}
To create our 450-tuple set of experimental materials, we sampled from \atypicnorm. To help narrow down the data, we decided to limit our experiments to context events that happen relatively quickly (take a full day or less to complete) and frequently in a person's life (at least once a year). So, via manual labeling, we set aside tuples that did not meet these criteria. However, we still consider these tuples to be high quality and hope that followup experiments can include them all.

\subsection{Dialogue Scenarios: Name selection} \label{subsec:appendix-names}
For both the dialogue names and the Mary/Linda name replacements, we choose to use names from non-English speaking countries that are rare in the United States. This is to fit with both our human and LLM experiments: we want to minimize the chance the LLMs have seen the exact sentences appear in their training data. We source names from a rare names list,\footnote{\url{https://github.com/sigpwned/popular-names-by-country-dataset}} dropping names from English-speaking countries and removing those which were in the top 10,000 names from a US census dataset.\footnote{\url{https://github.com/dxdc/babynames/blob/main/data/all-names.csv}} We balance names by gender and sample from 198 of them.
Note that we use the name `Anahita' in \cref{fig:example}, but the actual name \atypic is `Anohito.' Both names are valid spellings of a popular Tajik name; however, `Anohito' can easily be mixed up with a Japanese pronoun. So, to prevent confusion, we use `Anahita.'

\subsection{More explanation of the snarky label} \label{subsec:appendix-snarky-more}
When a speaker wants to communicate something negative about another person, they may communicate via inference so their on-record utterance is positive (``Hey, guess what, she meant it'') instead of saying something negative directly (``She usually doesn't mean it when she says thank you.''). 
As an example of a non-snarky dialogue, consider
``Anahita thanked her boss. And hey, guess what, she was in her boss's office!''. The atypicality inference (=Anahita usually does not thank her boss while in her boss's office) does not imply anything negative about Anahita -- maybe her and her boss usually meet outside of that office, but this does not make either of them look bad in any way. Hence, we would label this utterance as {\it not snarky}.

\paragraph{Relationship with commonsense knowledge \& irony}
In the context of our research questions, we can define snark as a specific way atypicality inferences can interact with commonsense knowledge to produce irony.

\textbf{Commonsense knowledge}: In the snark case, the \rwtypic event is not only generally typical, but is also an event that \textit{should} happen according to some social convention or standard of competence.

\textbf{Irony}: Verbal irony can be defined as a speaker expressing something by using language that would usually convey the opposite \cite{haverkate1990speech}, where the speaker expects the listener to recognize the ``critical or mocking attitude'' behind their words \cite{wilson2006pragmatics}. 
\citet{sperber1986relevance} have connected this ``critical or mocking attitude'' to  utterances which violate listeners' expectations of informativity, e.g. in our case with a speaker mentioning a (Context, \rwtypic) event pair.
In this paper, we call this ``critical or mocking attitude'' \textbf{Snark}.
After making an atypicality inference, a listener would only interpret the speaker as being snarky if the speaker's utterance was about someone fulfilling a social convention or meeting a minimal standard.
e.g., the atypicality inference from ``Anahita thanked her boss. And hey, guess what, she meant it!'' conveys that Anahita is usually the opposite of thankful.
In other words, if the atypicality inference can be interpreted as ironic, we label it as snarky.

\subsection{Pragmatics Pre-test for Human Participants}
\label{appendix:pretest}
\input{figs/pretest_impreq}
\input{figs/pretest_coh}
\paragraph{Justification and Source data}
We choose to focus on participants who we know are capable of making pragmatic inferences. Therefore, before recruiting participants for this experiment, we conduct an 18-question pragmatics pre-test using data from \texttt{PragMega}, a human-written multiple-choice dataset which includes questions regarding multiple pragmatic phenomena \cite{hu-etal-2023-fine, floyd2025three}.

\citet{floyd2025three} conclude that pragmatic reasoning can be broken into the understanding of 1) social conventions (SocCon), 2) intonation, and 3) world-knowledge based causal reasoning. Since the task of understanding atypicality inferences from text-only data requires both 1) and 3), we choose questions from those two categories: 12 SocCon (half implicatures, half implicit requests) and 6 commonsense-knowledge based coherence questions (half coherent, half incoherent). Half of our SocCon questions have literal answers as the correct answer, and the other half nonliteral. We release our pre-test questions to the public.\footnote{Withheld due to anonymity.} 

\paragraph{Usage and survey setup}
\cref{fig:pretest-impreq} and \cref{fig:pretest-coh} contain examples for an implicit request and coherence questions respectively.
We set up the pre-test on Prolific and conduct three separate rounds, where the pay is 
\$18.00
an hour averaged over the rounds. We give a passing score if a participant gets at least 5 out of 6 answers correct for each group of 6 questions (implicatures, requests, coherence) and finishes the test in 7 minutes or less. We do NOT inform participants that this is a pre-test, and we pay all participants regardless of their score. Out of 1,255 total participants, 1,196 are approved, 53 return, and 6 time out. Among the approved participants, 676 pass (56.52\%) and are subsequently invited to the main experiment.

\subsection{Ordinal Regressions} \label{subsec:appendix-ordinal}
We use mixed-effects ordinal regressions to model the rating category (very unlikely to very likely) with fixed effects for Follow-up type (\rwtypic/\rwatypic) and setting (\nullcond/Plain/HeyGW) and their interaction, along with random effects for Context utterance, and, in the human data case, participant. The mixed-effects ordinal regressions were built in R, using \texttt{clmm} from the library \texttt{ordinal}. For all regressions, fixed effects were centered and maximal random effect structure was included, as permitted by the data.

We predict opposite effects of Follow-up setting for \rwtypic (where an atypicality inference should yield lower ratings for Plain or HeyGW compared to \nullcond) and \rwatypic (where a typicality inference should yield lower ratings for \nullcond). We test the Snark effect in a further \textit{$\text{Follow-up} \times \text{setting} \times \text{snark}$} model, where the atypicality inference pattern is predicted to be stronger for snarky scenarios. 

\input{tables/human_ordinalreg_main}
\input{tables/human_ordinalreg_snark}
\input{tables/human_ordinal_subsets}
\paragraph{On human responses}
We collect 13,500 rating responses.
\cref{table:human-ordinal-main} reports the regression output for the main model (\textit{$\text{Follow-up} \times \text{setting}$}), and \cref{table:human-ordinal-snark} reports the output for the larger model with snark (\textit{$\text{Follow-up} \times \text{setting} \times \text{snark}$}). The table uses abbreviations for T(ypicality) and S(nark).  All interactions in both tables are significant, inviting the critical subset comparisons listed in \cref{table:human-ordinal-subsets}.  

\cref{table:human-ordinal-subsets} shows the subsequent analyses that test for a main effect of \textit{setting} in each of the relevant subsets. Regular font indicates a typicality inference, and bold font indicates an atypicality inference.

\input{tables/llm-R-table}
\paragraph{On LLM responses} 
In the same way as for the human data but without the participant ID random effect, we fit mixed effects models for each of the sets of LLM results. Like for humans, we fit models on the full LLM data. \cref{table:llm-r-results} summarizes the results. Only CoT-Prompting model configurations consistently follow human-like patterns, 4 out of the 5 making atypicality inferences, typicality inferences, and exhibiting the snark effect.

\input{figs/gpt5.4-shady}
\input{figs/llm_passedkappa_vs_infscore}
\subsection{Additional Figures} 
\label{subsec:appendix-figs}
As an example, \cref{fig:llm-shady-plot} shows the Snark effect for \texttt{GPT-5.4} (with CoT), but full results are in \cref{table:llm-r-results}.

\cref{fig:llm-passkapa-vs-infscore} shows a significant correlation ($R^2 = 0.235, p < 0.01$) between higher commonsense knowledge and more event-typicality inferences, which is in line with our belief that event-typicality inferences rely on commonsense knowledge.
To get the correlation values, we fit an ordinary least-squares linear model in R with the commonsense knowledge agreement as a predictor of InfScore ($\text{InfScore} {\sim} \kappa$).
\input{figs/llm-typinf-orig-flip}
\input{figs/llm-atypinf-orig_flip}

\section{Additional Discussion}
\label{appendix:discussion}
Here, we further discuss LLMs' oversensitivity to prompt wording instead of actual event typicality.
Because we found that LLMs and humans do not share the same Commonsense Knowledge, we wondered whether flipping the labels of each event in \atypicnorm to match the LLMs' voted response typicality (flipping from \rwtypic to \rwatypic and vice versa) would increase their number of ``correct'' inferences (i.e. typicality inferences for \rwatypic events and atypicality inferences for \rwtypic events). However, 
\cref{fig:llm-typinf-orig-flip} and \cref{fig:llm-atypinf-orig-flip} show that this does not work: the scores remain about the same whether we use the human knowledge labels or the LLM ones.

This is not surprising given the lack of consistency illustrated in \cref{fig:llm-norm-vs-none}. If the LLMs are responding to each question based on how it's asked (with dialogue context vs. without) and not based on any underlying knowledge about the events, then changing the event labels would not make a difference.

\section{Metric definitions} \label{sec:appendix-metrics-details}
We denote the dataset as $\mathcal{D}$ with $\vert\mathcal{D}\vert=2,354$ for \atypic
and $\vert \mathcal{D} \vert = 450$ for \atypicnorm.
Let $\mathcal{F} \in \{\text{`TypF'},\text{`AtypF'}\}$ be the Follow-up (abbreviated from \rwtypic and \rwatypic in order to save space). We collect $K$ responses for each datapoint $i$ and Follow-up $\mathcal{F}$. We use $R^{\mathcal{F}}_{i}[k]$ to indicate the $k$-th response ($k=1,\ldots,K$) to the question from $i$-th datapoint ($i=1,\ldots,\vert\mathcal{D}\vert$).

Each response $R^{\mathcal{F}}_{i}[k]$  takes one of the integer values from $\{1,2,3,4\}$, which map to the labels `Very Unlikely', `Unlikely', `Likely', and `Very Likely', respectively. For example, $R^{\text{AtypF}}_{i}[k]=3$ means that the $k$-th response to the $i$-th datapoint with a Follow-up of AtypF is `Likely'. We use two different ways of aggregating responses, which are detailed below.

\textbf{Average-response} $\mu^{\mathcal{F}}_{i} \in [1,4]$ is average of $K$ response values, \textit{i.e.},
\[
    \mu^{\mathcal{F},}_i = \frac{1}{K}\sum_{k=1}^{K}R^{\mathcal{F}}_{i}[k].
\]

\textbf{Voted-response} $v^{\mathcal{F}}_{i} \in \{-1,1\}$ is result of majority-voting from $K$ responses, specifically,
\[
    v^{\mathcal{F}}_{i}=
    \begin{cases}
        1 &\!\!\!\text{if }\! \operatorname{vote}^{\mathcal{F}}_i > \frac{1}{2} \\ 
        1 &\!\!\!\text{if }\! \left(\operatorname{vote}^{\mathcal{F}}_i = \frac{1}{2} \right) \!\text{ and }\! \left(\mu^{\mathcal{F}}_i \geq 2.5\right)\\
        -1 &\!\!\!\text{otherwise}
    \end{cases}
\]
where $\operatorname{vote}^{\mathcal{F}}_i = \frac{1}{K}\sum_{k=1}^{K}\mathbb{I}\left(R^{\mathcal{F}}_{i}[k]\in\{3,4\}\right)$. In words, $v^{\mathcal{F}}_{i}$ becomes $1$ (`Likely') if the majority of responses are positive (3 or 4), and $-1$ (`Unlikely') if the majority are negative (1 or 2). When a 50:50 tie occurs (\textit{i.e.}, $\operatorname{vote}^{\mathcal{F}}_i = \frac{1}{2}$), we break the tie using the average response. Let $\{v_i^{\text{TypF}}\}_{i=1}^{\vert\mathcal{D}\vert}$ be the set of all votes for \rwtypic{s} and  $\{v_i^{\text{AtypF}}\}_{i=1}^{\vert\mathcal{D}\vert}$ be the set of all votes for \rwatypic{s}. 

\textbf{Agreement $\kappa$}: We can now define the following three agreement scores, where $\kappa$ is Cohen's Kappa \cite{cohen1960coefficient}. Note we abbreviate \textbf{Combined} to \textbf{Comb} here for space.
\[
\begin{aligned}
\kappa^{\text{TypF}}   &= \kappa\!\left(\{v_i^{\text{TypF}}\}_{i=1}^{\lvert\mathcal{D}\rvert}\right), \\
\kappa^{\text{AtypF}}  &= \kappa\!\left(\{v_i^{\text{AtypF}}\}_{i=1}^{\lvert\mathcal{D}\rvert}\right), \\
\kappa^{\text{Comb}}
&= \kappa\!\left(
\left\{
\mathbf{1}\!\left[
v_i^{\text{TypF}} = 1 \wedge v_i^{\text{AtypF}} = -1
\right]
\right\}_{i=1}^{\lvert\mathcal{D}\rvert}
\right)
\end{aligned}
\]

Now, let $\mathcal{S} \in \{\text{`\nullcond', `Plain', `HeyGW'}\}$ be the Follow-up \textbf{setting} from our Event-Typicality Inference experiment (see \cref{table:exp-setup-materials}). All the definitions from above stay the same, but we additionally use $R^{\mathcal{F},\mathcal{S}}_{i}[k]$ (e.g. $R^{\text{AtypF}, \text{Plain}}_{i}[k]=3$ means the $k$-th response to the $i$-th datapoint with (AtypF, Plain)), along with $\mu^{\mathcal{F},\mathcal{S}}_i$ and $v^{\mathcal{F},\mathcal{S}}_i$.

\textbf{Consistency score} $C^{\mathcal{F}}\in[0,1]$, which is defined for each $\mathcal{F}$, is the proportion of datapoints from $\mathcal{D}$ where the voted-response in the Commonsense Knowledge experiment $v^{\mathcal{F}}_i$ matches the voted-response with \nullcond setting in the Event-Typicality Inference experiment $v^{\mathcal{F},\text{\nullcond}}_i$.  
\textit{i.e.},
\begin{align*}
C^{\text{TypF}}&=\frac{1}{\vert\mathcal{D}\vert}\sum_{i=1}^{\vert\mathcal{D}\vert}\mathbb{I}\left(v^{\text{TypF}}_i = v^{\text{TypF},\text{\nullcond}}_i \right),\\
C^{\text{AtypF}}&=\frac{1}{\vert\mathcal{D}\vert}\sum_{i=1}^{\vert\mathcal{D}\vert}\mathbb{I}\left(v^{\text{AtypF}}_i = v^{\text{AtypF},\text{\nullcond}}_i \right).
\end{align*}

\textbf{Inference score (\infscore)} $I^{\mathcal{F},\mathcal{S}_{\text{Mentioned}}}_i \in [-3,3]$, which is defined for each datapoint $i$ with some $\mathcal{F}$, measures the change in average-response when a MENTIONED setting (Plain or HeyGW) is introduced, compared to the baseline setting without followup (\nullcond), \textit{i.e.},
\[
I^{\mathcal{F},\mathcal{S}_{\text{Mentioned}}}_i = \mu^{\mathcal{F},\text{\nullcond}}_{i}-\mu^{\mathcal{F},\mathcal{S}_{\text{Mentioned}}}_{i},
\]
where $\mathcal{S}_{\text{Mentioned}}\in\{\text{`Plain'},\text{`HeyGW'}\}$. A positive \infscore indicates an atypicality inference, a negative \infscore indicates a typicality inference, and zero indicates that no inference occurred.

\section{All LLM-filtering prompts}
\label{sec:appendix-all-prompts}
\subsection{Abstract}
\input{prompts/llm_filtering/group/abstract/abstract-abstract_direct}
\input{prompts/llm_filtering/group/abstract/abstract-abstract_idiom}
\input{prompts/llm_filtering/group/abstract/abstract-abstract_imagine}

\subsection{Entailment or near-entailment} \label{subsec:entail}
\input{prompts/llm_filtering/group/entail/entail-entail_direct}

\input{prompts/llm_filtering/group/entail/entail-entail_death}
\input{prompts/llm_filtering/group/entail/entail-entail_definition}
\input{prompts/llm_filtering/group/entail/entail-entail_para}

\input{prompts/llm_filtering/group/entail/entail-entail_plausible}

\subsection{Not sufficiently likely} \label{subsec:appendix-expect}
Note that the prompt may look strange for \cref{prompt:expect-expect-direct}. This is the result of iterative prompt engineering to get the best result. Previously, more intuitive prompts resulted in the LLM rejecting almost everything or almost nothing.

\input{prompts/llm_filtering/group/expect/expect-expect_direct}
\input{prompts/llm_filtering/group/expect/expect-expect_extra}
\input{prompts/llm_filtering/group/expect/expect-expect_nondefault}

\subsection{Malformed context}
\input{prompts/llm_filtering/group/malformed-a/malformed_a-malformed_a_base}
\input{prompts/llm_filtering/group/malformed-a/malformed_a-malformed_a_redundant}
\subsection{Malformed event utterance}
\input{prompts/llm_filtering/group/malformed-ab/malformed_ab-malformed_ab_base}
\input{prompts/llm_filtering/group/malformed-ab/malformed_ab-malformed_ab_tense}
\input{prompts/llm_filtering/group/malformed-ab/malformed_ab-malformed_ab_coherent}

\subsection{Utterance not brief}
\input{prompts/llm_filtering/indiv/brevity}

\subsection{Strange use of definites}
\input{prompts/llm_filtering/indiv/definite}

\subsection{Narration}
\input{prompts/llm_filtering/indiv/narrative}

\subsection{Context event not sufficiently frequent}
\input{prompts/llm_filtering/group/life/life-dailylife}
\input{prompts/llm_filtering/group/life/life-lifetime}

\subsection{Time-gap}
\input{prompts/llm_filtering/indiv/gap}

\subsection{Time-scope}
\input{prompts/llm_filtering/indiv/scope}

\subsection{Passive}
\input{prompts/llm_filtering/indiv/passive}

\subsection{Not standalone}
\input{prompts/llm_filtering/indiv/standalone}

\subsection{Too vague}
\input{prompts/llm_filtering/indiv/vague}

\subsection{No typicality}

\input{prompts/llm_filtering/indiv/typicality_q}
\subsection{Prompts that had no effect in the end}

\input{prompts/llm_filtering/indiv/explain}

\input{prompts/llm_filtering/group/result/result-result_ab}

\input{prompts/llm_filtering/group/result/result-result_ba}
\input{prompts/llm_filtering/indiv/habitual}

%% file: tables/rwtypic-examples.tex
\begin{table}
  \begin{center}
    \small
    \begin{tabular}{l|l} 
      \toprule
    \textbf{\rwtypic Event} & \textbf{Example followup to} \\
     Category & Mary thanks her boss \\
    \midrule \midrule
    State & Mary is at the office \\
    \midrule
    Subaction & Mary means it \\
    \midrule
    Next-action & Mary's boss says you're welcome \\
      \bottomrule
    \end{tabular}
  \end{center}
      \caption{Example \rwtypic events for the three utterance categories.}
    \label{table:category-def}
\end{table}

%% file: prompts/initial_filtering/classify_easy.tex
\begin{lstlisting}[style=prompt,caption={classify-easy},label={prompt:classify-easy},captionpos=b]
You are an expert annotator of natural language. You will be presented with a clause.
Your task is to determine if any of the following is true:

- The clause is incomplete
- The clause is malformed
- The clause is abstract.
- The clause is idiomatic or not literal.
- The clause contains a passive.
- The clause contains an implicit verb.
- The clause does not take simple present tense form.
- The clause contains a pronoun with no antecedent.
- The clause contains a qualitative or evaluative action.
- The clause contains a habitual or stative verb form, or its meaning can be interpreted as habitual/stative.

If any of the above are true, respond with the tuple (True, [above item which most clearly is true]). Otherwise, respond with the tuple (False, None).

CLAUSE: Mary loses Linda arm
RESPONSE: (True, 'malformed')

CLAUSE: Mary loses her keys
RESPONSE: (False, None)

CLAUSE: Mary reinvents the wheel
RESPONSE: (True, 'idiomatic')

CLAUSE: Mary wants a car.
RESPONSE: (True, 'stative')

CLAUSE: Mary knows would die
RESPONSE: (True, 'malformed')

CLAUSE: Mary exposes oneself
RESPONSE: (True, 'pronoun')

CLAUSE: Mary sings to her
RESPONSE: (True, 'pronoun')

CLAUSE: Mary sings at the competition
RESPONSE: (False, None)

CLAUSE: Mary spends a day
RESPONSE: (True, 'incomplete')

CLAUSE: Mary buys some apples
RESPONSE: (False, None)

CLAUSE: Mary sings to me
RESPONSE: (True, 'pronoun')

CLAUSE: Mary exposes them
RESPONSE: (True, 'pronoun')

CLAUSE: Mary means nothing to Linda
RESPONSE: (True, 'abstract')

CLAUSE: Linda forgets her coat
RESPONSE: (True, 'implicit verb')

CLAUSE: Mary gets insulted by Linda
RESPONSE: (True, 'passive')

CLAUSE: Linda insults Mary
RESPONSE: (False, None)

CLAUSE: Mary insults him
RESPONSE: (True, 'pronoun')

CLAUSE: Mary is calm in stressful situations
RESPONSE: (True, 'habitual')

CLAUSE: Mary does well at the competition
RESPONSE: (True, 'evaluative')

CLAUSE: Mary competes against Linda
RESPONSE: (False, None)

CLAUSE: Mary gets on the bus
RESPONSE: (False, None)

CLAUSE: Mary remembers her coat
RESPONSE: (True, 'implicit verb')

CLAUSE: Mary wants to eat lunch
RESPONSE: (True, 'not simple present')

CLAUSE: Mary eats lunch
RESPONSE: (False, None)

CLAUSE: {}
RESPONSE: 
\end{lstlisting}

%% file: prompts/initial_filtering/classify_hard.tex
\begin{lstlisting}[style=prompt,caption={classify-hard},label={prompt:classify-hard},captionpos=b]
You are an expert annotator of natural language. You will be presented with a clause.
Your task is to determine if any of the following is true:

- The clause describes a vague action that would typically be seen with more context, e.g. with a modifier or with a more detailed predicate.- The clause appears to be grounded in a specific context that is not specified.
- The clause describes a situation which is extremely unusual in daily life. Perhaps the situation is common on fictional stories, but not real life.
- The clause describes an action which is abnormal or nonsensical without context.
If any of the above are true, respond with the tuple (True, [above item which most clearly is true]). Otherwise, respond with the tuple (False, None).

CLAUSE: Mary loses her keys
RESPONSE: (False, None)

CLAUSE: Mary buys some apples
RESPONSE: (False, None)

CLAUSE: Mary leaves the apple
RESPONSE: (True, 'needs context')

CLAUSE: Linda kills Mary
RESPONSE: (True, 'unusual')

CLAUSE: Mary sings at the competition
RESPONSE: (False, None)

CLAUSE: Mary comes over
RESPONSE: (True, 'vague')

CLAUSE: Mary comes over to Linda's house
RESPONSE: (False, None)

CLAUSE: Mary smells the nail
RESPONSE: (False, 'needs context')

CLAUSE: Mary smells the flower
RESPONSE: (True, None)

CLAUSE: Mary tells Linda
RESPONSE: (True, 'vague')

CLAUSE: Mary steals Linda's child
RESPONSE: (True, 'unusual')

CLAUSE: Mary competes against Linda
RESPONSE: (False, None)

CLAUSE: Mary beats her swords into ploughshares
RESPONSE: (True, 'abnormal')

CLAUSE: Linda hits the apple
RESPONSE: (True, 'abnormal')

CLAUSE: Mary eats lunch
RESPONSE: (False, None)

CLAUSE: Mary gets on the bus
RESPONSE: (False, None)

CLAUSE: Mary finds out her biological dad is not hers
RESPONSE: (True, 'unusual')

CLAUSE: {}
RESPONSE: 
\end{lstlisting}

%% file: prompts/rwtypic_gen/gen3clauses.tex
\begin{lstlisting}[style=prompt,caption={gen3clause},label={prompt:gen3clause},captionpos=b]
You are an expert annotator of natural language. You will be presented with a clause, which is presumably about taking an action or being at a specific place or event.
Generate three potential next clauses: A state of being, a sub-action, and a subsequent action.

Criteria for all clauses:
- The clause must be about a natural assumption given the original clause which feels obvious and redundant.
- The clause must not be a direct entailment of the original clause
- The clause must not be a paraphrase of the original clause
- The clause can be about the same entity as that in the original, or about another entity associated with the original.
- The clause you generate should be brief. Do not add extra modifiers or details to it.

Criteria for the state of being clause:
- The state of being must not be a presupposition of the original clause.
- It must not be directly inferable from the original clause. In other words, there cannot be a logical entailment relationship.

Criteria for the sub-action clause:
- It must not be a direct entailment of the original clause.
- It must not be a paraphrase of the same action in the original clause.
- It might not be possible to generate a good sub-action clause. For example, when the original clause contains no action, or contains an action that is too short to contain a sub-action, it may not be possible.
Criteria for the subsequent action clause:
- It must be a very obvious assumption given the original clause, to the point that it seems redundant to explicitly mention in natural speech.
- It should be as general or vague as possible, with no details that would make it unique or important to mention in real life.
Your response format should be a three-item tuple of (state of being, sub-action, subsequent action). Each item in the tuple is optional. When there is no good clause for that category, respond with None in that index of the tuple. If there are no good clauses to generate at all, respond with (None, None, None).CLAUSE: Mary goes to the restaurant
RESPONSE: ("Mary is hungry", None, "Mary eats")
CLAUSE: Mary goes to a picnic
RESPONSE: ("Mary wants to", "Mary brings a blanket", "Mary eats")
CLAUSE: Mary sings at the contest
RESPONSE: (None, "The crowd notices", "Mary receives applause")
CLAUSE: Linda buys candy at the store
RESPONSE: ("Linda has money on her", "Linda pays the cashier", "Linda does not throw the candy away")
CLAUSE: Linda attends a concert
RESPONSE: ("Linda knows the band", "Linda sings along", "Linda claps at the end")
CLAUSE: Mary attends the talk
RESPONSE: ("Mary knows who the speaker is", "Mary pays attention", "Mary does not regret it")
CLAUSE: Mary drives home after a long day of work
RESPONSE: (None, "Mary has both hands on the wheel", "Mary parks without trouble")
CLAUSE: Mary waves her hands in Linda's face
RESPONSE: ("Mary knows Linda", None, "Linda notices")
CLAUSE: Mary does the reading for next week's math class
RESPONSE: ("Mary is enrolled in the class", "Mary absorbs some of it", "Mary attends the next class")
CLAUSE: {}
RESPONSE: 
\end{lstlisting}

%% file: tables/rejection_twoclass.tex
\begin{table*}[t]
\centering
\resizebox{\textwidth}{!}{%
\begin{tabular}{@{}ll|ll}
\toprule
 \multicolumn{2}{c}{Unnatural for dialogue}   &  \multicolumn{2}{c}{Wrong typicality}   \\
\midrule
   \textbf{Prompt(s)} &
  \textbf{Description} & 
    \textbf{Prompt(s)} &
  \textbf{Description} \\
\midrule 
 Malformed Context (2): & Has errors or is & Abstract (3): & Idiomatic, not concrete, or solely \\
\cref{prompt:malformed-a-malformed-a-base}, \cref{prompt:malformed-a-malformed-a-redundant}
  & awkward/redundant &  \cref{prompt:abstract-abstract-direct},
  \cref{prompt:abstract-abstract-idiom}, \cref{prompt:abstract-abstract-imagine} &  happens in one person's head \\
\midrule
 Malformed event utterance (3): & Has errors or & Entailed/almost entailed (5): &   \\
 \cref{prompt:malformed-ab-malformed-ab-base},
\cref{prompt:malformed-ab-malformed-ab-tense},
\cref{prompt:malformed-ab-malformed-ab-coherent}
& is incoherent & Direct (\cref{prompt:entail-entail-direct} & \\
 &  & Death (\cref{prompt:entail-entail-death}), & Event only false if the person dies  \\
&  & Definition (\cref{prompt:entail-entail-definition}), & Event is definition of context  \\
&  & Paraphrase (\cref{prompt:entail-entail-para}), & Event is paraphrase of context  \\
&  & Plausibility (\cref{prompt:entail-entail-plausible}) & Implausible that it didn't happen  \\
\midrule
Not brief (\cref{prompt:brevity}) & Event utterance & Not sufficiently likely (3):  &   \\
& is too descriptive & Unexpected (\cref{prompt:expect-expect-direct}), & Not RW-Typic \\
  &  & Extra info (\cref{prompt:expect-expect-extra}), & Adds info/detail \\
   &  & Non-default (\cref{prompt:expect-expect-nondefault}) & Interprets context strangely  \\
\midrule
Should be definite (\cref{prompt:definite}) & `the' needed & Context event not sufficiently &   \\ 
& & frequent (\cref{prompt:life-dailylife}, \cref{prompt:life-lifetime}) &   \\ 
\midrule
Narration (\cref{prompt:narrative}) & Context-Event pair reads & Time-gap (\cref{prompt:gap}) & Context and event utterances  \\ 
  & like text, not dialogue &  & take place far apart in time  \\ 
  \midrule
 Passive  (\cref{prompt:passive}) & Context utterance in & Time-scope (\cref{prompt:scope}) & Context and event utterances \\ 
      & passive voice &  & are not in the same time-scope  \\ 
\cmidrule(l){3-4}
      &  & Not standalone (\cref{prompt:standalone}) &  Context utterance needs more \\ 
    &  &  & context to stand alone \\ 
\cmidrule(l){3-4}
    &  & Too vague (\cref{prompt:vague}) & Underspecified, uninterpretable \\ 
\cmidrule(l){3-4}
    &  & No typicality (\cref{prompt:typicality-q}) & E.g. Event utterance is \\ 
     &  & & about an inherent trait \\
\bottomrule
\end{tabular}
}
\caption{27 LLM-filtering prompts and their classes (Unnatural for dialogue vs. Wrong typicality)} \label{table:llm-filter-prompt-classes} 
\end{table*}

%% file: prompts/llm_filtering/gen-often-q.tex
\begin{lstlisting}[style=prompt,caption={Prompt used to generate the ``When...how often'' typicality question for the LLM-filtering.},label={prompt:gen-often-q},captionpos=b]
You are an expert annotator of natural language data.
You will read STATEMENT_ONE and STATEMENT_TWO.
Your task is to produce a QUESTION in the format of "When [STATEMENT_ONE], how often [STATEMENT_TWO]?
When STATEMENT_ONE uses 'the' to modify its object, please switch to using 'a' in the first clause of your generated QUESTION.
But if STATEMENT_TWO uses 'the' to modify its object, please use 'the' in the second clause of the QUESTION.

STATEMENT_ONE: When Mary finds the stray dog
STATEMENT_TWO: Mary pets it
QUESTION: When Mary finds a stray dog, how often does she pet it?
STATEMENT_ONE: When Mary finds a stray dog
STATEMENT_TWO: Mary is an animal lover
QUESTION: When Mary finds a stray dog, how often is she an animal lover?
STATEMENT_ONE: Mary builds the house
STATEMENT_TWO: Mary takes a whole year
QUESTION: When Mary builds a house, how often does she take a whole year?
STATEMENT_ONE: Mary builds Linda's house
STATEMENT_TWO: Linda moves in
QUESTION: When Mary builds Linda's house, how often does Linda move in?
STATEMENT_ONE: Mary makes cookies
STATEMENT_TWO: Mary takes the cookies to the office
QUESTION: When Mary makes cookies, how often does she take the cookies to the office?
STATEMENT_ONE: Mary makes friendship bracelets
STATEMENT_TWO: Mary uses more than one bead shape
QUESTION: When Mary makes friendship bracelets, how often does she use more than one bead shape?
STATEMENT_ONE: Mary makes friendship bracelets
STATEMENT_TWO: Mary enjoys crafting as a hobby
QUESTION: When Mary makes friendship bracelets, how often does she enjoy crafting as a hobby?
STATEMENT_ONE: Mary sees shoes for sale
STATEMENT_TWO: Mary buys them
QUESTION: When Mary sees shoes for sale, how often does she buy them?
STATEMENT_ONE: Mary sees shoes for sale
STATEMENT_TWO: Mary is interested in fashion
QUESTION: When Mary sees shoes for sale, how often is she interested in fashion?
STATEMENT_ONE: {}
STATEMENT_TWO: {}
QUESTION: 
\end{lstlisting}

%% file: prompts/rwatypic_gen/gen-rwatypic-state.tex
\begin{lstlisting}[style=prompt,caption={Prompt to generate the \rwatypic state events.},label={prompt:gen-rwatypic-state},captionpos=b]
You will be presented with two utterances, ORIGINAL and STATE. They appear in a sequence, where ORIGINAL is what a person does, and STATE describes a corresponding state of being.
Your task is to generate your version of a STATE utterance (STATE_YOURS).
STATE_YOURS should be temporary: that is, it cannot be something that usually lasts for weeks, months, or years.
Examples of some non-temporary states to avoid generating: Mary is a woman, Mary has a car, Mary has a disability.
The most notable exception to this rule is the state of knowing. For example, in 'Mary knows the band,' the concept of knowing a band can be applied to many other bands, so it can apply to multiple situations in a person's life.
In addition to being temporary in a person's life, make sure STATE_YOURS is more specific than STATE. It also must be as close to the original STATE length as possible. It is acceptable to make it just one word longer, one word shorter, or the same length.

ORIGINAL: Mary goes to the restaurant
STATE: Mary is hungry
STATE_YOURS: Mary is ravenous
ORIGINAL: Mary goes to a picnic
STATE: Mary wants to
NEXT_YOURS: Mary is starving
ORIGINAL: Mary sings at the contest
STATE: Mary sounds good
STATE_YOURS: Mary sounds magnificent
ORIGINAL: Linda buys candy at the store
STATE: Linda has money on her
STATE_YOURS: Linda has a craving
ORIGINAL: Linda attends a concert
STATE: Linda knows the band
STATE_YOURS: Linda knows every song
ORIGINAL: Mary attends the talk
STATE: Mary knows who the speaker is
STATE_YOURS: Mary knows the speaker's mentor
ORIGINAL: Mary drives home after a long day of work
STATE: Mary has her hands on the wheel
STATE_YOURS: Mary has plans to watch a movie
ORIGINAL: Mary does the reading for next week's math class
STATE: Mary is enrolled in the class
STATE_YOURS: Mary is bored with it
ORIGINAL: {}
STATE: {}
STATE_YOURS: 
\end{lstlisting}

%% file: prompts/rwatypic_gen/gen-rwatypic-subaction.tex
\begin{lstlisting}[style=prompt,caption={Prompt to generate the \rwatypic subaction events.},label={prompt:gen-rwatypic-subaction},captionpos=b]
You will be presented with two utterances, ORIGINAL and SUBACTION. They appear in a sequence, where ORIGINAL is what a person does, and SUBACTION describes a corresponding subaction.
Your task is to generate your version of a SUBACTION utterance (SUBACTION_YOURS). Make it more interesting than SUBACTION.
But try to keep it as close to the original SUBACTION length as possible. It is acceptable if you make it just one word longer, one word shorter, or the same length.

ORIGINAL: Mary goes to the restaurant
SUBACTION: Mary drives there
SUBACTION_YOURS: Mary runs late
ORIGINAL: Mary goes to a picnic
SUBACTION: Mary brings a blanket
SUBACTION_YOURS: Mary brings mayonnaise
ORIGINAL: Mary sings at the contest
SUBACTION: The crowd notices
SUBACTION_YOURS: The crowd goes wild
ORIGINAL: Linda buys candy at the store
SUBACTION: Linda pays the cashier
SUBACTION_YOURS: Linda uses coupons
ORIGINAL: Linda attends a concert
SUBACTION: Linda sings along
SUBACTION_YOURS: Linda screams along
ORIGINAL: Mary attends the talk
SUBACTION: Mary pays attention
SUBACTION_YOURS: Mary falls asleep
ORIGINAL: Mary drives home after a long day of work
SUBACTION: Mary keeps her hands on the wheel
SUBACTION_YOURS: Mary gets caught in traffic
ORIGINAL: Mary waves her hands in Linda's face
SUBACTION: Linda notices it
SUBACTION_YOURS: Linda ignores it
ORIGINAL: Mary does the reading for next week's math class
SUBACTION: Mary absorbs some of it
SUBACTION_YOURS: Mary absorbs nothing
ORIGINAL: {}
SUBACTION: {}
SUBACTION_YOURS: 
\end{lstlisting}

%% file: prompts/rwatypic_gen/gen-rwatypic-next.tex
\begin{lstlisting}[style=prompt,caption={Prompt to generate the \rwatypic next-events.},label={prompt:gen-rwatypic-next},captionpos=b]
You will be presented with two utterances, ORIGINAL and NEXT. They appear in a sequence, where ORIGINAL is what a person does, and NEXT is what happens next.
Your task is to generate your version of a NEXT utterance (NEXT_YOURS). Make it more specific than NEXT.
But try to keep it as close to the original NEXT length as possible. It's best if you don't make it any longer, but it is acceptable if you make it just one word longer.

ORIGINAL: Mary goes to the restaurant
NEXT: Mary eats food
NEXT_YOURS: Mary eats pizza
ORIGINAL: Mary goes to a picnic
NEXT: Mary eats food
NEXT_YOURS: Mary eats two sandwiches
ORIGINAL: Mary sings at the contest
NEXT: The audience claps at the end
NEXT_YOURS: Mary gets third place
ORIGINAL: Linda buys candy at the store
NEXT: Linda pays the cashier
NEXT_YOURS: Linda gives some to Mary
ORIGINAL: Linda attends a concert
NEXT: Linda leaves when it ends
NEXT_YOURS: Linda leaves late
ORIGINAL: Mary attends the talk
NEXT: Mary does not regret it
NEXT_YOURS: Mary talks to the speaker
ORIGINAL: Mary drives home after a long day of work
NEXT: Mary parks without trouble
NEXT_YOURS: Mary heats up leftovers
ORIGINAL: Mary waves her hands in Linda's face
NEXT: Linda notices it
NEXT_YOURS: Linda is annoyed
ORIGINAL: Mary does the reading for next week's math class
NEXT: Mary attends the next math class
NEXT_YOURS: Mary barely understands it
ORIGINAL: {}
NEXT: {}
NEXT_YOURS: 
\end{lstlisting}

%% file: tables/attention-checks.tex
\begin{table*}
  \begin{center}
    \small
    \resizebox{\textwidth}{!}{
    \begin{tabular}{l|lll}
      \toprule 
    & \textbf{Accepted Answer} & \textbf{Norming version} & \textbf{Main version}  \\
    \midrule \midrule 
    1 & Likely/Very Likely & 
    In general, when someone puts on 
    & Bob and Sarah are having a conversation     \\
    & & a t-shirt and skirt, how likely is & about their friend Sharon. Bob: \\
    & & it that they put clothes on? & ``Sharon put on a t-shirt and skirt. \\
    & & & She put clothes on.'' In general, \\
    & & & when Sharon puts on a t-shirt and skirt, \\
    & & &  how likely is it that she puts clothes on? \\
    \hdashline
    2 & Likely/Very Likely & 
    In general, when someone plays a  
    & Linda and Sabrina are having a conversation \\
    & & game of chess, how likely is & about their friend Grace. Linda: \\
    & & it that they play a game? & ``Grace played a game of chess. \\
    & & & She played a game.'' In general, \\
    & & & when Grace plays a game of chess, \\
    & & &  how likely is it that she plays a game? \\
    \hdashline
    3 & Likely/Very Likely  & 
    In general, when someone enters  
    & "Simon and Jared are having a conversation \\
    & & a REM cycle, how likely is & about their friend David. Simon: \\
    & & it that they are asleep? & ``David entered a REM cycle. He was \\
    & & & asleep.'' In general, when \\
    & & & David enters a REM cycle, \\
    & & &  how likely is it that he is asleep? \\
    \midrule
    4 & Unlikely/Very Unlikely  & 
    In general, when someone goes to  
    & George and Sally are having a conversation      \\
    & & a grocery store, how likely is it that & about their friend Jane. George: \\
    & & they give birth inside? & ``Jane went to the grocery store. \\
    & & &  She gave birth inside.'' In general, when \\
    & & & Jane goes to a grocery store, how \\
    & & & likely is it that she gives birth inside? \\
    \hdashline
    5 & Unlikely/Very Unlikely & 
    In general, when someone works a 
    & Mildred and Lily are having a conversation  \\
    & & shift at the restaurant, how likely is it that  & about their friend Steve. Mildred: \\
    & & they call an ambulance for a customer? & ``Steve worked a shift at the restaurant. \\
    & & &  He called an ambulance for a customer.'' \\
    & & &  In general, when Steve works a shift at \\
    & & &  the restaurant, how likely is it that \\
    & & &  he calls an ambulance for a customer? \\
    \hdashline
    6 & Unlikely/Very Unlikely & 
    In general, when someone sees a dog, 
    & Jack and Stacy are having a conversation   \\
    & & how likely is it that it bites off & about their friend Joseph. Jack: \\
    & & their left hand? & ``Joseph saw the dog. It bit off \\
    & & & his left hand.'' In general, when \\
    & & & Joseph sees a dog, how likely is it \\
    & & & that it bites off his left hand? \\
      \bottomrule 
    \end{tabular}
    }
  \end{center}
      \caption{Attention check questions. These qualify as `nonsensical items' per Prolific's attention check policy.}
    \label{table:attention-checks}
\end{table*}

%% file: tables/all_llms.tex
\begin{table*}[t]
\centering
\resizebox{0.7\textwidth}{!}{%
\begin{tabular}{@{}l|llll}
\toprule
LLM & Base & Instruct & Small & Big \\
\midrule 
GPT-5.4 \citep{openai2026gpt54systemcard} & x & \checkmark & \textit{nano-2026-03-17} & \textit{2026-03-05}  \\
Llama-3.1 \citep{grattafiori2024llama} & \checkmark & \checkmark & 8b & 70b \\
Olmo-3 \citep{olmo2025olmo} & \checkmark &  \checkmark & 7b & 32b \\
Qwen3 \citep{yang2025qwen3} & \checkmark & \checkmark & 8b & 30b \\ 
gemma-3 \citep{gemmateam2025gemma3technicalreport}  & \checkmark & \checkmark & 12b & 27b  \\
\bottomrule
\end{tabular}
}
\caption{LLMs used. For those where prompting was available, we tried prompting both with and without CoT (see \cref{subsubsec:prompting-details}. For details on model version selection, see \cref{subsubsec:olmo-qwen}. Since OpenAI has not made GPT-5.4's parameter number publicly available, we just put the timestamp model IDs.} \label{table:llms-used} 
\end{table*}

%% file: prompts/prompting/without-reasoning.tex
\begin{lstlisting}[style=prompt,caption={An example prompt used for the prompting experiment \textit{without} CoT and an example response (from Llama-3.1-70B).},label={prompt:no-CoT-response},captionpos=b]
[Prompt] Chedeline and Kathem are having a conversation about their friend Anahita. Chedeline: "Anahita thanked her boss. And hey, guess what, she was in tears!" In general, when Anahita thanks her boss, how likely is it that she is in tears?\nChoose one of the following options: Very Unlikely, Unlikely, Likely, Very Likely.

[Response] "Unlikely"
\end{lstlisting}

%% file: prompts/prompting/with-reasoning.tex
\begin{lstlisting}[style=prompt,
caption={An example prompt used for the prompting experiment \textit{with} CoT and an example response (from Llama-3.1-70B).},label={prompt:CoT-response},captionpos=b]
[Prompt] Chedeline and Kathem are having a conversation about their friend Anahita. Chedeline: "Anahita thanked her boss. And hey, guess what, she meant it!" In general, when Anahita thanks her boss, how likely is it that she means it?\nAnswer the question with the following schema: {"type": "object", "properties": {"reasoning": {"type": "string", "description": "The step-by-step reasoning process for the answer."}, "answer": {"type": "string", "enum": ["Very Unlikely", "Unlikely", "Likely", "Very Likely"]}}, "required": ["reasoning", "answer"]}

[Response] "{\"reasoning\": \"The statement is that Anahita thanked her boss and actually meant it. The tone of the conversation suggests surprise, implying that it is not typical for Anahita to genuinely thank her boss.\", \"answer\": \"Very Unlikely\"}"
\end{lstlisting}

%% file: prompts/norming/gen-norm-anon-q.tex
\begin{lstlisting}[style=prompt,caption={Prompt used to generate the typicality questions used in the norming study.},label={prompt:gen-norm-anon-q},captionpos=b]
You are an expert annotator of natural language data.
You will read STATEMENT_ONE and STATEMENT_TWO.
Your task is to produce a QUESTION in the format of "When [STATEMENT_ONE], how likely is it that [STATEMENT_TWO]?
When STATEMENT_ONE uses 'the' to modify its object, please switch to using 'a' in the first clause of your generated QUESTION.
But if STATEMENT_TWO uses 'the' to modify its object, please use 'the' in the second clause of the QUESTION.
If only one person exists in the statements, use the word 'someone' instead of their name.
If two people exist, use the phrase 'one person' instead of the first name and 'second person' for the second name.

STATEMENT_ONE: Mary finds the stray dog
STATEMENT_TWO: Mary pets it
QUESTION: When someone finds a stray dog, how likely is it that they pet it?
STATEMENT_ONE: Mary finds a stray dog
STATEMENT_TWO: Mary is happy
QUESTION: When someone finds a stray dog, how likely is it that they are happy?
STATEMENT_ONE: Mary builds the house
STATEMENT_TWO: Mary takes a whole year
QUESTION: When someone builds a house, how likely is it that they take a whole year?
STATEMENT_ONE: Mary builds Linda's house
STATEMENT_TWO: Linda moves in
QUESTION: When one person builds a second person's house, how likely is it that the second person moves in?
STATEMENT_ONE: Mary makes cookies
STATEMENT_TWO: Mary takes the cookies to the office
QUESTION: When someone makes cookies, how likely is it that they take the cookies to the office?
STATEMENT_ONE: Mary makes friendship bracelets
STATEMENT_TWO: Mary uses more than one bead shape
QUESTION: When someone makes friendship bracelets, how likely is it that they use more than one bead shape?
STATEMENT_ONE: Mary cooks pasta for Linda
STATEMENT_TWO: Linda eats it
QUESTION: When one person cooks pasta for a second person, how likely is it that the second person eats it?
STATEMENT_ONE: {}
STATEMENT_TWO: {}
QUESTION: 
\end{lstlisting}

%% file: figs/norm_screenshot.tex
\begin{figure*}[t]
\centering
\includegraphics[width=0.6\textwidth]{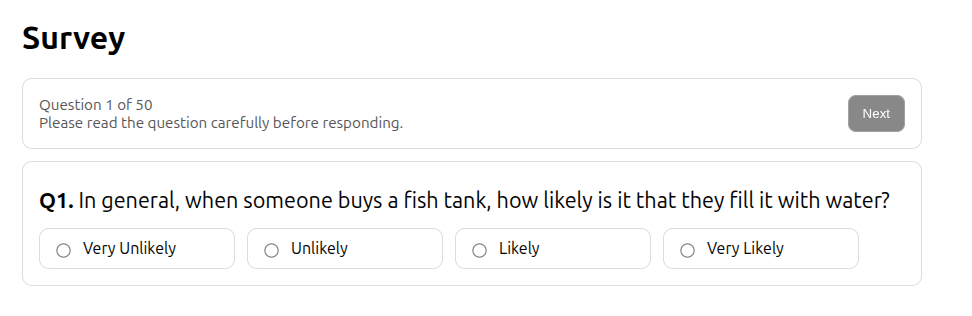}
\caption{Example of what human participants saw in our Commonsense Knowledge experiment.}
\label{fig:norm-ui}
\end{figure*}

%% file: figs/mainexp_screenshot.tex
\begin{figure*}[t]
\centering
\includegraphics[width=0.6\textwidth]{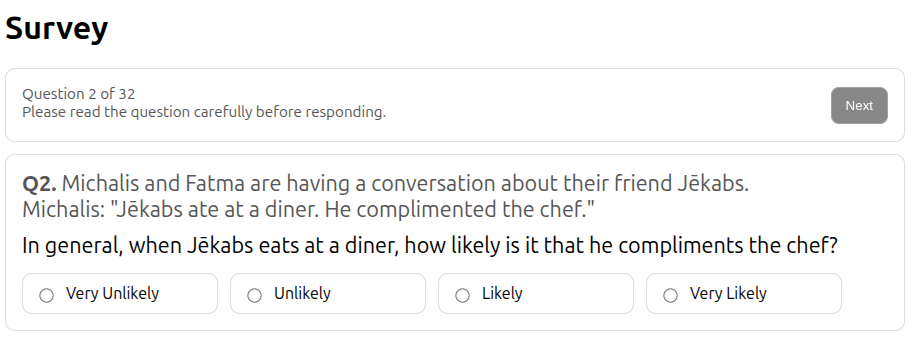}
\caption{Example of what human participants saw during our Event-Typicality Inference experiment. Each participant answered between 30 and 32 questions.}
\label{fig:mainexp-screenshot}
\end{figure*}

%% file: figs/pretest_impreq.tex
\begin{figure*}[t]
\centering
\includegraphics[width=0.6\textwidth]{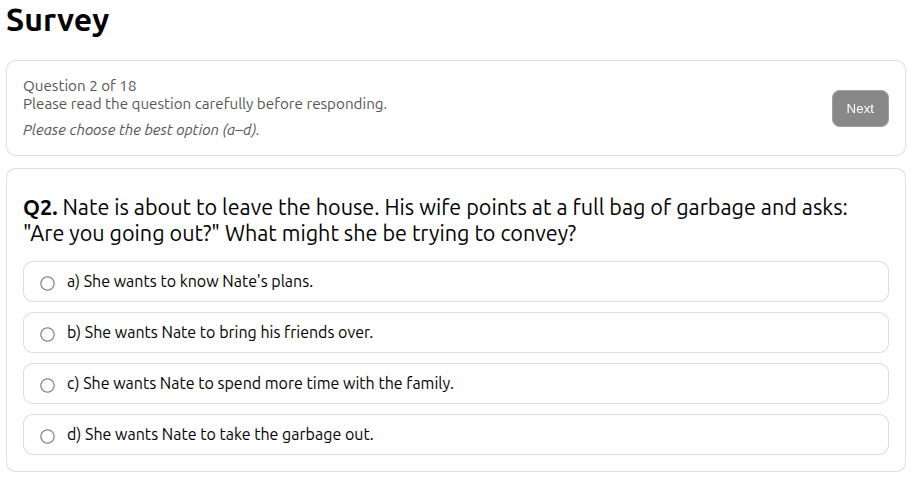}
\caption{Example of an implicit request question participants saw in the pragmatics pre-test. The expected answer here is d).}
\label{fig:pretest-impreq}
\end{figure*}

%% file: figs/pretest_coh.tex
\begin{figure*}[t]
\centering
\includegraphics[width=0.6\textwidth]{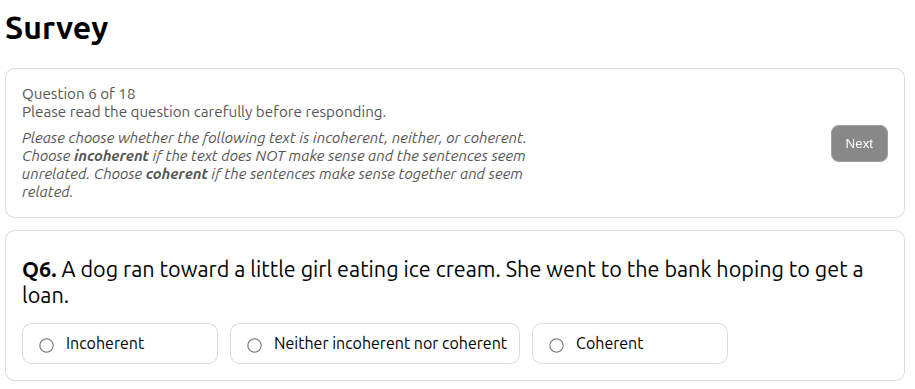}
\caption{Example of a coherence question participants saw in the pragmatics pre-test. The expected answer here is `Incoherent.'}
\label{fig:pretest-coh}
\end{figure*}

%% file: tables/human_ordinalreg_main.tex
\begin{table}[t]
\centering
\resizebox{\columnwidth}{!}{%
\begin{tabular}{llll}
\toprule
  &	$\beta$		& z	& p\\
\midrule
	Follow-up          & 2.87 & 39.85&  $<$0.001\\
	Plain.\nullcond & 1.20 & 26.15&  $<$0.001\\
	\nullcond.HeyGW & -0.27 & -5.28&  $<$0.001\\
    Follow-up:Plain.\nullcond & -1.05 & -11.47&  $<$0.001\\
    Follow-up:\nullcond.HeyGW & 1.95 & 18.51&  $<$0.001\\
\bottomrule
\end{tabular}
}
\caption{Regression output for the main mixed effects model fit on human data (\textit{$\text{Follow-up} \times \text{setting}$}).}
\label{table:human-ordinal-main}
\end{table}

%% file: tables/human_ordinalreg_snark.tex
\begin{table}[t]
\centering
\resizebox{\columnwidth}{!}{
\begin{tabular}{llll}
\toprule
  &	$\beta$		& z	& p\\
\midrule
	Follow-up             & 2.77& 40.57& $<$0.001\\
	Plain.\nullcond     & 1.18& 26.50& $<$0.001\\
	\nullcond.HeyGW     & -0.24& -4.72& $<$0.001\\
	Snark              & -0.22& -3.46& $<$0.001\\
	Follow-up:Plain.\nullcond   & -1.02& -11.88& $<$0.001\\
	Follow-up:\nullcond.HGW   & 1.95& 22.69& $<$0.001\\
	Follow-up:Snark            & -0.87& -6.83& $<$0.001\\
	Plain.\nullcond:Snark   & -0.005&-0.06 & $<$0.001\\
	\nullcond.HeyGW:Snark   & 0.25& 2.53& $<$0.001\\
	Follow-up:Plain.\nullcond:Snark & -0.04& -0.24& $<$0.001\\
	Follow-up:\nullcond.HeyGW:Snark & 0.15& 0.92& $<$0.001\\
\bottomrule
\end{tabular}
}
\caption{Regression output for the snark mixed effects model fit on human data, \textit{$\text{Follow-up} \times \text{setting} \times \text{snark}$}.}
\label{table:human-ordinal-snark}
\end{table}

%% file: tables/human_ordinal_subsets.tex
\begin{table}[t]
\centering
\resizebox{0.7\columnwidth}{!}{%
\begin{tabular}{llll}
  &	$\beta$		& z	& p\\
\toprule
\multicolumn{4}{c}{\rwtypic}\\
\midrule
Plain.\nullcond	& 0.66	&  10.06  & $<$0.001\\
{\bf{\nullcond.HeyGW}}	& {\bf{0.64}}	& {\bf{8.77}}	&{\bf{$<$0.001}}\\
\midrule
\multicolumn{4}{c}{\rwatypic}\\
\midrule
Plain.\nullcond	& 1.86	& 26.50	& $<$0.001\\
\nullcond.HeyGW	& -1.34	& -17.98	& $<$0.001\\
\midrule
\multicolumn{4}{c}{{\rwtypic}-Snark}\\
\midrule
Plain.\nullcond	& 0.65	& 7.06	&  $<$0.001\\
{\bf{\nullcond.HeyGW}}	& {\bf{0.81}}	& 	{\bf{7.17}}& {\bf{$<$0.001}}\\
\midrule
\multicolumn{4}{c}{{\rwtypic}-NoSnark}\\
\midrule
Plain.\nullcond	& 0.65	& 7.08	&  $<$0.001\\
{\bf{\nullcond.HeyGW}}	& {\bf{0.54}}	& 	{\bf{5.64}}& {\bf{$<$0.001}}\\
\midrule
\multicolumn{4}{c}{{\rwatypic}-Snark}\\
\midrule
Plain.\nullcond	& 1.87	& 17.01	& $<$0.001\\
\nullcond.HeyGW	& -1.24	& -11.13	& $<$0.001\\
\midrule
\multicolumn{4}{c}{{\rwatypic}-NoSnark}\\
\midrule
Plain.\nullcond	& 1.85	& 	20.51& $<$0.001\\
\nullcond.HeyGW	& -1.43	& 	-14.15& $<$0.001\\
\bottomrule
\end{tabular}

}
\caption{Regression output for the subset mixed effects models on human data.}
\label{table:human-ordinal-subsets}
\end{table}

%% file: tables/llm-R-table.tex
\begin{table*}[t]
\centering
\resizebox{\textwidth}{!}{%
\begin{tabular}{@{}l|ccc|ccc|ccc}
\toprule
LLM
& \multicolumn{3}{c}{Surprisal}   
&  \multicolumn{3}{c}{Prompting} 
& \multicolumn{3}{c}{CoT-Prompting}   \\
\cmidrule(lr){2-10}
& AtypInf & TypInf & SnarkEffect 
& AtypInf & TypInf & SnarkEffect 
& AtypInf & TypInf & SnarkEffect \\
\midrule 
GPT-5.4 & -- & -- & -- & x & \checkmark & x & \checkmark & \checkmark & \checkmark \\
Llama-3.1-70B & \checkmark & x & \checkmark & \checkmark & \checkmark & \checkmark & \checkmark & \checkmark & \checkmark \\
Olmo3-32B & \checkmark & x & x & x & \checkmark & x & \checkmark & \checkmark & x \\
Qwen3-30B & x & \checkmark & x & x & \checkmark & x & x & \checkmark & x \\
gemma-3-27b & x & x & x & x & \checkmark & x & \checkmark & \checkmark & \checkmark \\
\midrule \midrule
GPT-5.4-nano & -- & -- & -- & x & \checkmark & x & x & \checkmark & x \\
Llama-3.1-8B & \checkmark & x & x & x & \checkmark & x & \checkmark & \checkmark & \checkmark \\
Olmo3-7B & x & \checkmark & x & x & \checkmark & x & x & \checkmark & x \\
Qwen3-8B & \checkmark & x & \checkmark & x & \checkmark & x & x & \checkmark & x \\
gemma-3-12b & x & x & x & x & \checkmark & x & x & \checkmark & x \\
\bottomrule
\end{tabular}
}
\caption{Summary of mixed effects models fit on LLM output data. Atypicality and typicality inferences abbreviated to AtypInf and TypInf. Note that the snark effect is only present if atypicality inferences are made.} \label{table:llm-r-results} 
\end{table*}

%% file: figs/gpt5.4-shady.tex
\begin{figure}
    \centering
    \includegraphics[width=\linewidth]{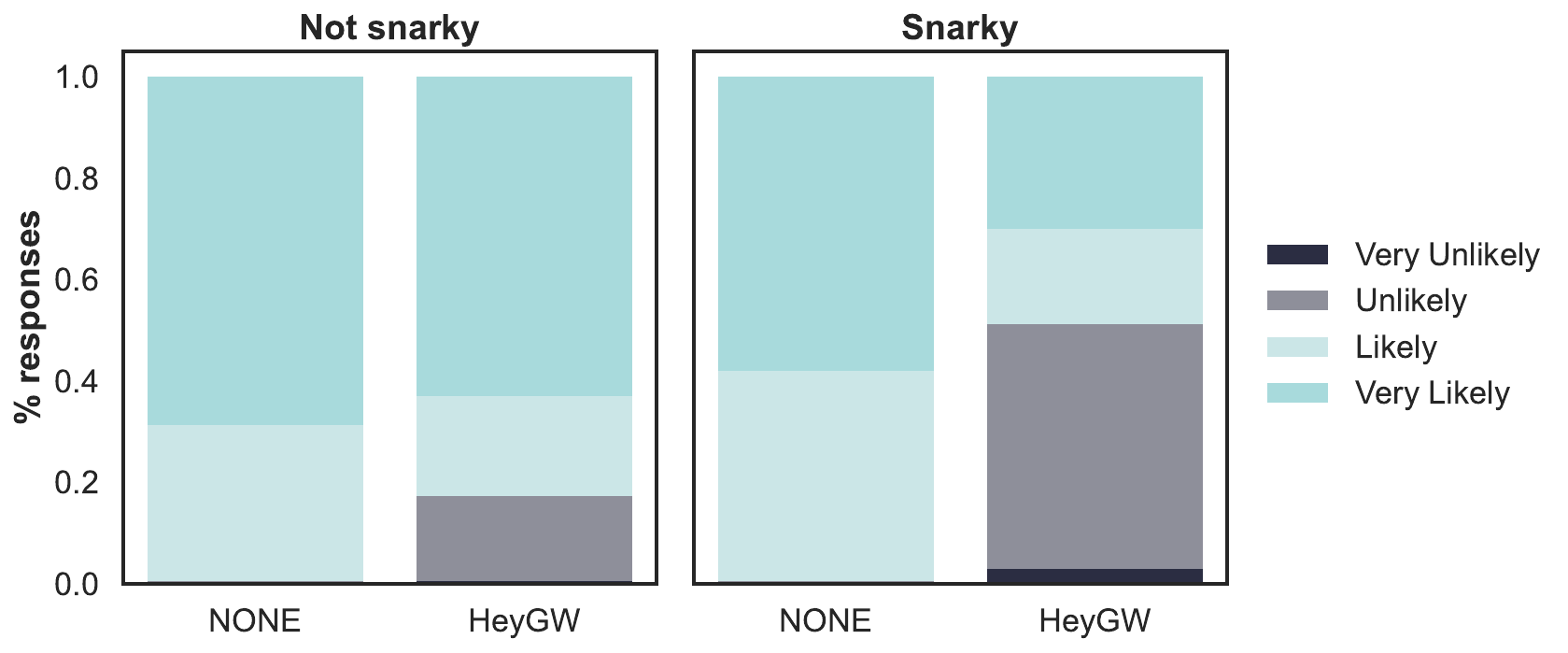}
    \caption{\texttt{GPT-5.4} comparison of the distribution of responses to \nullcond vs. HeyGW settings for \rwtypic{s}, with non-snarky dialogues on the left and snarky dialogues on the right.}
    \vspace*{-0.3em}
    \label{fig:llm-shady-plot}
    \vspace*{-0.7em}
\end{figure}

%% file: figs/llm_passedkappa_vs_infscore.tex
\begin{figure}[t]
\centering
\includegraphics[width=\columnwidth]{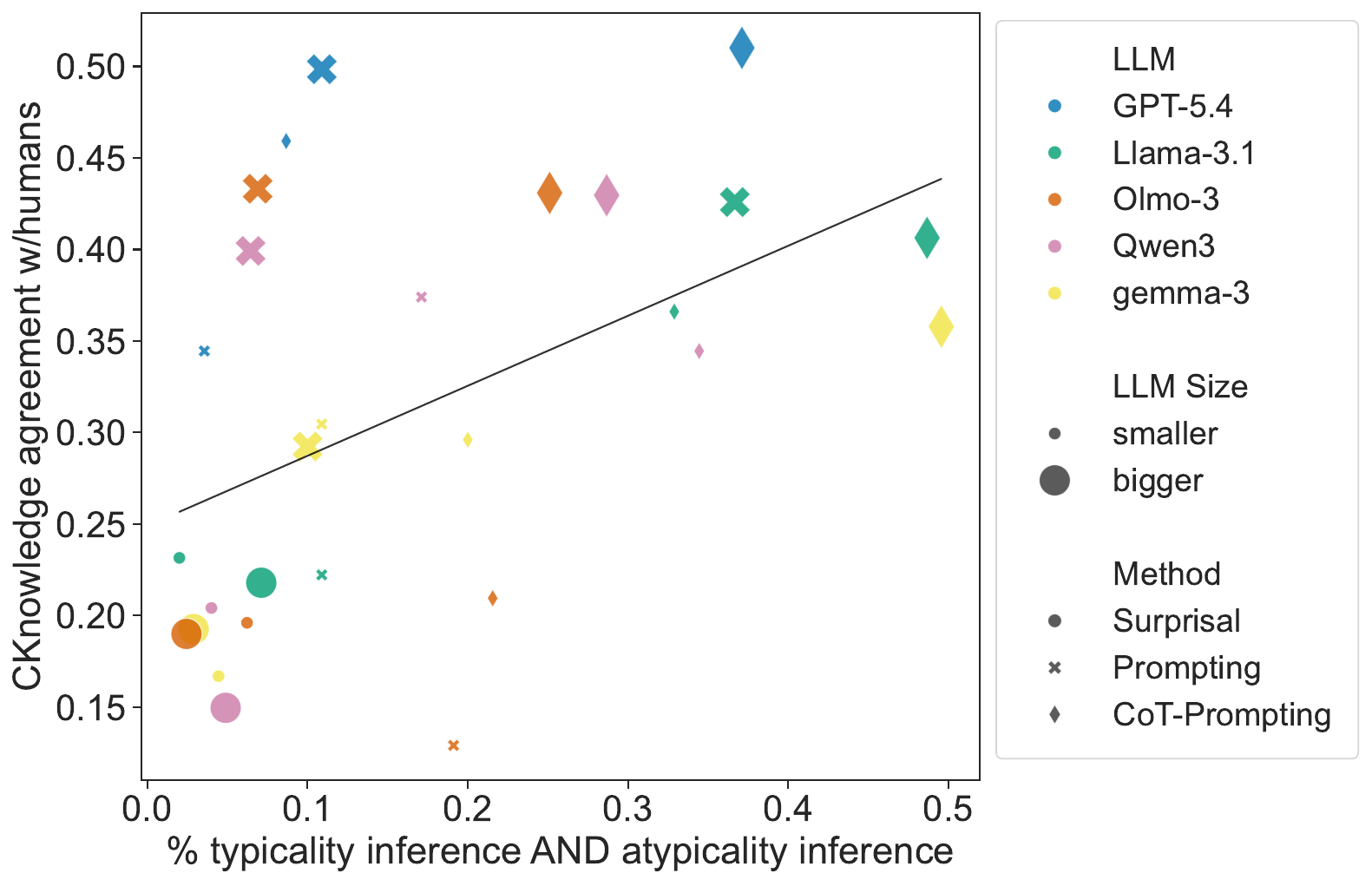}
\caption{Correlation (p<0.01) between Commonsense Knowledge (CKnowledge) human-LLM agreement ($\kappa^\text{Combined}$, \cref{sec:appendix-metrics-details}) and the percentage of Event-Typicality Inference Experiment HeyGW setting tuples
where the LLM made both an atypicality inference (on the \rwtypic) \textit{and} a typicality inference (on the \rwatypic).}
\label{fig:llm-passkapa-vs-infscore}
\vspace*{-0.7em}
\end{figure}

%% file: figs/llm-typinf-orig-flip.tex
\begin{figure}[t]
\centering
\includegraphics[width=\columnwidth]{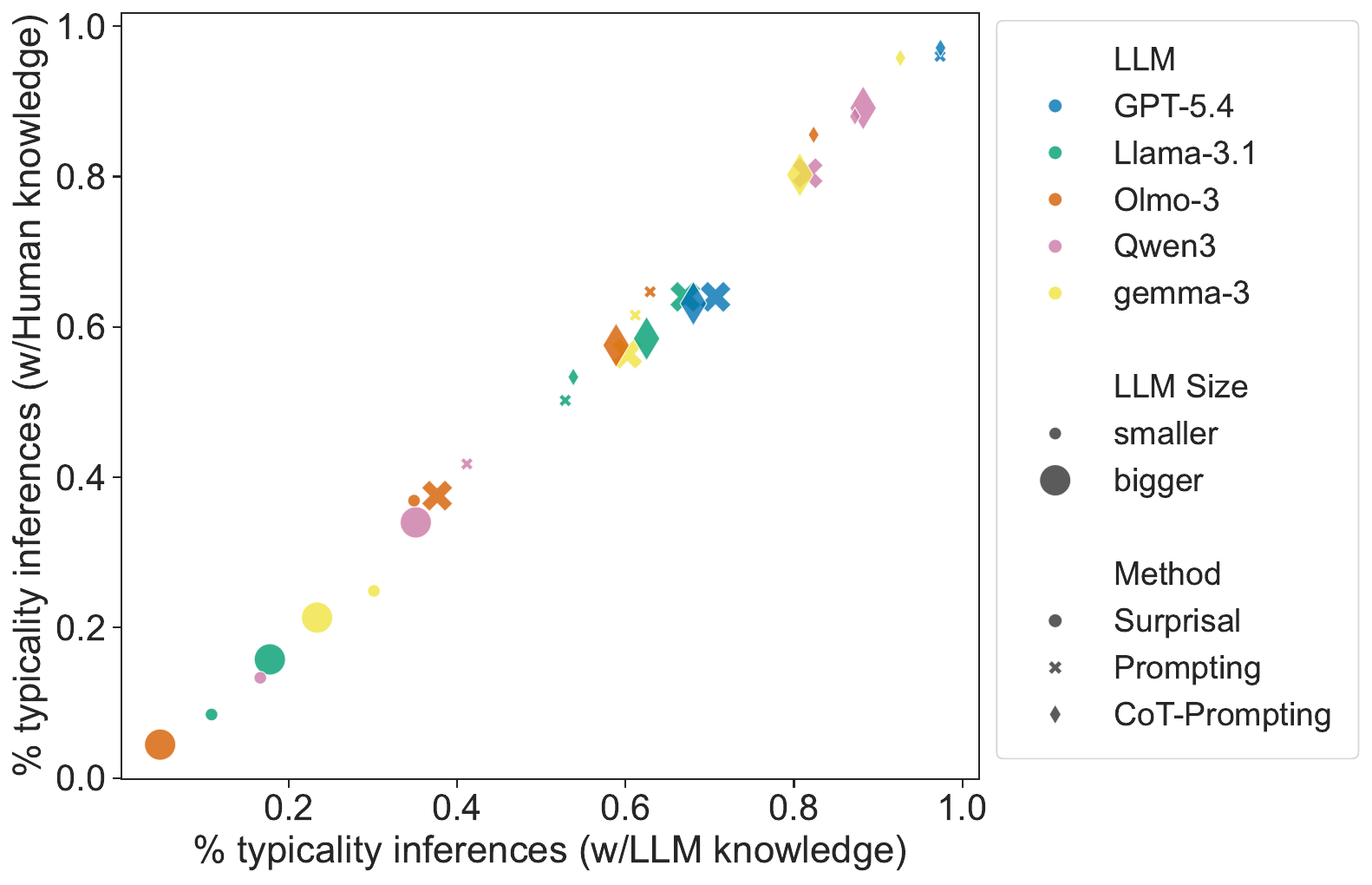}
\caption{Percentage of the time a typicality inference was made on events voted as \rwatypic by LLM (X-axis) vs. humans (Y-Axis)}
\label{fig:llm-typinf-orig-flip}
\vspace*{-0.7em}
\end{figure}

%% file: figs/llm-atypinf-orig_flip.tex
\begin{figure}[t]
\centering
\includegraphics[width=\columnwidth]{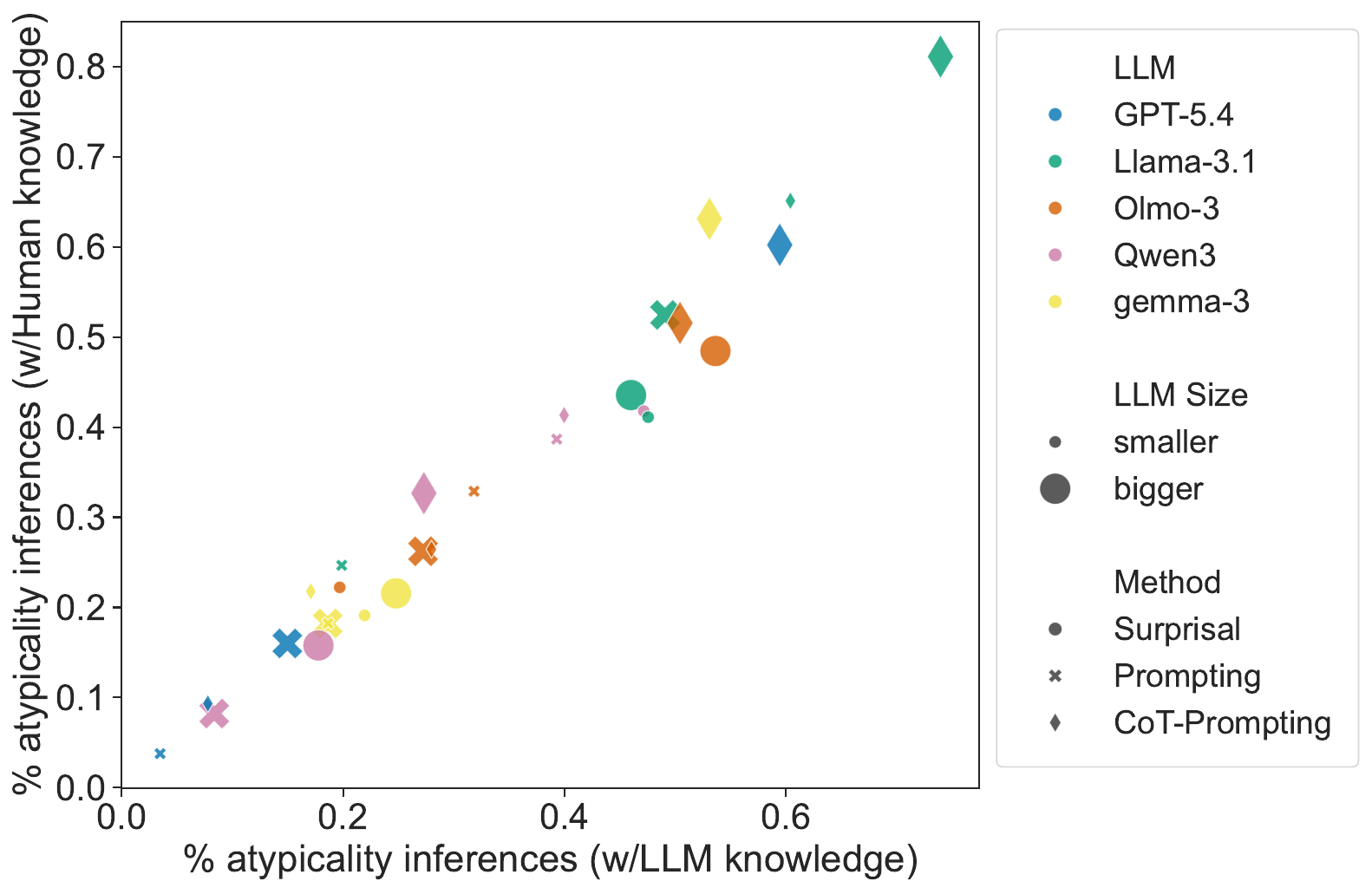}
\caption{Percentage of the time an atypicality inference was made on events voted as \rwtypic by LLM (X-axis) vs. humans (Y-Axis)}
\label{fig:llm-atypinf-orig-flip}
\vspace*{-0.7em}
\end{figure}

%% file: prompts/llm_filtering/group/abstract/abstract-abstract_direct.tex
\begin{lstlisting}[style=prompt,caption={Abstract-Direct: Applies to only Context utterances.},label={prompt:abstract-abstract-direct},captionpos=b]
You are an expert annotator of natural language data.
You will read a QUESTION, a STATEMENT, and some EXAMPLES.
To the best of your knowledge, and according to the examples, answer the QUESTION with yes or no.

QUESTION: Is this statement about an abstract action? Definition of an abstract action: an action that may encompass a series of undefined, highly contextual, and potentially abstract steps.
STATEMENT: Mary loses hope
ANSWER: Yes
STATEMENT: Mary loses the bet
ANSWER: No
STATEMENT: Mary asserts her independence
ANSWER: Yes
STATEMENT: Mary gets Linda a job
ANSWER: Yes
STATEMENT: Mary gets Linda a glass of water
ANSWER: No
STATEMENT: Mary drives Linda crazy
ANSWER: Yes
STATEMENT: Mary drives home
ANSWER: No
STATEMENT: Mary applies herself
ANSWER: Yes
STATEMENT: Mary applies for a job
ANSWER: No
STATEMENT: {Context}
ANSWER: 
\end{lstlisting}

%% file: prompts/llm_filtering/group/abstract/abstract-abstract_idiom.tex
\begin{lstlisting}[style=prompt,caption={Abstract-Idiom: Applies to only Context utterances.},label={prompt:abstract-abstract-idiom},captionpos=b]
You are an expert annotator of natural language data.
You will read a QUESTION, a STATEMENT, and some EXAMPLES.
To the best of your knowledge, and according to the examples, answer the QUESTION with yes or no.

QUESTION: Does this statement contain any idioms?
STATEMENT: Mary reinvents the wheel
ANSWER: Yes
STATEMENT: Mary eats an apple
ANSWER: No
STATEMENT: {Context}
ANSWER: 
\end{lstlisting}

%% file: prompts/llm_filtering/group/abstract/abstract-abstract_imagine.tex
\begin{lstlisting}[style=prompt,caption={Abstract-Imagine: Applies to only Context utterances.},label={prompt:abstract-abstract-imagine},captionpos=b]
You are an expert annotator of natural language data.
You will read a QUESTION, a STATEMENT, and some EXAMPLES.
To the best of your knowledge, and according to the examples, answer the QUESTION with yes or no.

QUESTION: Does what is expressed in this statement happen solely in someone's head? As in, it's their thought or their feeling?
STATEMENT: Mary feels a sharp sting
ANSWER: Yes
STATEMENT: Mary realizes Linda's error
ANSWER: Yes
STATEMENT: Mary considers buying them
ANSWER: Yes
STATEMENT: Mary jumps
ANSWER: No
STATEMENT: {Context}
ANSWER: 
\end{lstlisting}

%% file: prompts/llm_filtering/group/entail/entail-entail_direct.tex
\begin{lstlisting}[style=prompt,caption={Entail-Direct: Applies to all Context-Event pairs.},label={prompt:entail-entail-direct},captionpos=b]
You are an expert annotator of natural language data.
You will read a QUESTION, STATEMENT_A, STATEMENT_B, and some EXAMPLES.
To the best of your knowledge, and according to the examples, answer the QUESTION with yes or no.

QUESTION: Does statement A entail statement B?
STATEMENT_A: Mary makes a reservation
STATEMENT_B: Mary provides a time
ANSWER: Yes
STATEMENT_A: Mary makes a reservation
STATEMENT_B: Mary plans to go
ANSWER: No
STATEMENT_A: Mary invites Linda to come to her party
STATEMENT_B: Mary is having a party
ANSWER: Yes
STATEMENT_A: Mary invites Linda to come to her party
STATEMENT_B: Linda accepts the invitation
ANSWER: No
STATEMENT_A: {Context}
STATEMENT_B: {Event}
ANSWER: 
\end{lstlisting}

%% file: prompts/llm_filtering/group/entail/entail-entail_death.tex
\begin{lstlisting}[style=prompt,caption={Entail-Death: Applies to all Context-Event pairs.},label={prompt:entail-entail-death},captionpos=b]
You are an expert annotator of natural language data.
You will read a QUESTION, STATEMENT_A, STATEMENT_B, and some EXAMPLES.
To the best of your knowledge, and according to the examples, answer the QUESTION with yes or no.

QUESTION: Is death or the apocalypse the only way you can imagine statement B being false? - Falling asleep \rightarrow Waking up,  - Going somewhere \rightarrow Leaving, - Doing an action that always has some kind of end \rightarrow It ending
STATEMENT_A: Mary visits Linda's father
STATEMENT_B: Mary leaves
ANSWER: Yes
STATEMENT_A: Mary visits Linda's father
STATEMENT_B: Linda's father recognizes her
ANSWER: No
STATEMENT_A: Mary takes a hike
STATEMENT_B: Mary returns home
ANSWER: Yes
STATEMENT_A: Mary takes a hike
STATEMENT_B: Mary follows the trail
ANSWER: No
STATEMENT_A: Mary goes to sleep
STATEMENT_B: Mary wakes up
ANSWER: Yes
STATEMENT_A: Mary covers her head
STATEMENT_B: Mary uncovers her head later
ANSWER: Yes
STATEMENT_A: {Context}
STATEMENT_B: {Event}
ANSWER: 
\end{lstlisting}

%% file: prompts/llm_filtering/group/entail/entail-entail_definition.tex
\begin{lstlisting}[style=prompt,caption={Entail-Definition: Applies to all Context-Event pairs.},label={prompt:entail-entail-definition},captionpos=b]
You are an expert annotator of natural language data.
You will read a QUESTION, STATEMENT_A, STATEMENT_B, and some EXAMPLES.
To the best of your knowledge, and according to the examples, answer the QUESTION with yes or no.

QUESTION: Can the main point of statement B be seen as a part of the definition of an action in statement A? Note that if, according to its definition, statement A can happen without statement B occurring, your answer should be No.
STATEMENT_A: Mary plays chess
STATEMENT_B: Mary makes a move
ANSWER: Yes
STATEMENT_A: Mary plays chess
STATEMENT_B: Mary wins
ANSWER: No
STATEMENT_A: Mary buys groceries
STATEMENT_B: Mary selects food items
ANSWER: Yes
STATEMENT_A: Mary buys groceries
STATEMENT_B: Mary selects apples
ANSWER: No
STATEMENT_A: Mary follows Linda's steps
STATEMENT_B: Mary reaches the same destination as Linda
ANSWER: Yes
STATEMENT_A: {Context}
STATEMENT_B: {Event}
ANSWER: 
\end{lstlisting}

%% file: prompts/llm_filtering/group/entail/entail-entail_para.tex
\begin{lstlisting}[style=prompt,caption={Entail-Para: Applies to all Context-Event pairs.},label={prompt:entail-entail-para},captionpos=b]
You are an expert annotator of natural language data.
You will read a QUESTION, STATEMENT_A, STATEMENT_B, and some EXAMPLES.
To the best of your knowledge, and according to the examples, answer the QUESTION with yes or no.

QUESTION: Can statement B be seen as a paraphrase of statement A?
STATEMENT_A: Mary gets a ticket
STATEMENT_B: Mary takes the ticket
ANSWER: Yes
STATEMENT_A: Mary gets a ticket
STATEMENT_B: Mary goes to the event
ANSWER: No
STATEMENT_A: {Context}
STATEMENT_B: {Event}
ANSWER: 
\end{lstlisting}

%% file: prompts/llm_filtering/group/entail/entail-entail_plausible.tex
\begin{lstlisting}[style=prompt,caption={Entail-Plausible: Applies to all Context-Event pairs.},label={prompt:entail-entail-plausible},captionpos=b]
You are an expert annotator of natural language data.
You will read a QUESTION, STATEMENT_A, STATEMENT_B, and some EXAMPLES.
To the best of your knowledge, and according to the examples, answer the QUESTION with yes or no.

QUESTION: Is it extremely implausible for statement B to be false given statement A? In other words, does it take mental gymnastics to imagine statement B being false?
STATEMENT_A: Mary invites Linda to join
STATEMENT_B: Mary has plans
ANSWER: Yes
STATEMENT_A: Mary invites Linda to join
STATEMENT_B: Linda accepts the invitation
ANSWER: No
STATEMENT_A: Mary makes chocolate chip cookies
STATEMENT_B: Mary mixes the ingredients
ANSWER: Yes
STATEMENT_A: Mary makes chocolate chip cookies
STATEMENT_B: Mary doubles the recipe
ANSWER: No
STATEMENT_A: Mary accuses Linda of cheating
STATEMENT_B: Linda responds
ANSWER: Yes
STATEMENT_A: Mary accuses Linda of cheating
STATEMENT_B: Linda denies it
ANSWER: No
STATEMENT_A: Mary watches the film
STATEMENT_B: The credits roll
ANSWER: Yes
STATEMENT_A: {Context}
STATEMENT_B: {Event}
ANSWER: 
\end{lstlisting}

%% file: prompts/llm_filtering/group/expect/expect-expect_direct.tex
\begin{lstlisting}[style=prompt,caption={Expect-Direct: Applies to all Context-Event pairs.},label={prompt:expect-expect-direct},captionpos=b]
You are an expert annotator of natural language data.
You will read a QUESTION, STATEMENT_A, STATEMENT_B, and some EXAMPLES.
To the best of your knowledge, and according to the examples, answer the QUESTION with yes or no.

QUESTION: Given statement A, how surprised are you that statement B is true? On a scale of 0 to 10, where 10 is the most surprised, if your surprisal is 4 or less, answer Yes. Otherwise, answer No.
STATEMENT_A: Mary gives Linda a small smile
STATEMENT_B: Linda acknowledges it
ANSWER: Yes
STATEMENT_A: Mary gives Linda a small smile
STATEMENT_B: Mary is in a joyful mood
ANSWER: No
STATEMENT_A: Mary picks berries
STATEMENT_B: Mary takes them home
ANSWER: Yes
STATEMENT_A: Mary picks berries
STATEMENT_B: Mary is in a forest
ANSWER: No
STATEMENT_A: Mary decides to try out for the team
STATEMENT_B: Mary plans to attend practice
ANSWER: Yes
STATEMENT_A: Mary decides to try out for the team
STATEMENT_B: Mary fills out an application
ANSWER: No
STATEMENT_A: Mary gets Linda a job
STATEMENT_B: Mary thanks Linda
ANSWER: Yes
STATEMENT_A: Mary gets Linda a job
STATEMENT_B: Mary loves Linda
ANSWER: No
STATEMENT_A: Mary catches the dog
STATEMENT_B: Mary makes sure it gets home
ANSWER: Yes
STATEMENT_A: Mary catches the dog
STATEMENT_B: Mary clings to the dog
ANSWER: No
STATEMENT_A: {Context}
STATEMENT_B: {Event}
ANSWER: 
\end{lstlisting}

%% file: prompts/llm_filtering/group/expect/expect-expect_extra.tex
\begin{lstlisting}[style=prompt,caption={Expect-Extra: Applies to all Context-Event pairs.},label={prompt:expect-expect-extra},captionpos=b]
You are an expert annotator of natural language data.
You will read a QUESTION, STATEMENT_A, STATEMENT_B, and some EXAMPLES.
To the best of your knowledge, and according to the examples, answer the QUESTION with yes or no.

QUESTION: Does statement B add unexpected information that is not contained in statement A? The most common way of adding unexpected information is the addition of extra entities or characters into the situation. In the prize example, you can see 'judges' being added, where they did not exist and were not automatically inferable from statement A. In the drink example, some vague person who needs an apology has been added.
STATEMENT_A: Mary wins the prize
STATEMENT_B: Mary thanks the judges
ANSWER: Yes
STATEMENT_A: Mary wins the prize
STATEMENT_B: Mary is happy
ANSWER: No
STATEMENT_A: Mary spills a drink
STATEMENT_B: Mary apologizes
ANSWER: Yes
STATEMENT_A: Mary spills a drink
STATEMENT_B: Mary cleans it up
ANSWER: No
STATEMENT_A: {Context}
STATEMENT_B: {Event}
ANSWER: 
\end{lstlisting}

%% file: prompts/llm_filtering/group/expect/expect-expect_nondefault.tex
\begin{lstlisting}[style=prompt,caption={Expect-Nondefault: Applies to all Context-Event pairs.},label={prompt:expect-expect-nondefault},captionpos=b]
You are an expert annotator of natural language data.
You will read a QUESTION, STATEMENT_A, STATEMENT_B, and some EXAMPLES.
To the best of your knowledge, and according to the examples, answer the QUESTION with yes or no.

QUESTION: Does statement B interpret statement A in a non-default way? The most common way to take a photo is to use a smartphone, in which case storing the photo and adjusting the focus should happen automatically. So, statement B seems to be interpreting statement A as being about a real camera, which is non-default.
STATEMENT_A: Mary takes a photo
STATEMENT_B: Mary stores the photo
ANSWER: Yes
STATEMENT_A: Mary takes a photo
STATEMENT_B: Mary adjusts the focus
ANSWER: Yes
STATEMENT_A: Mary takes a photo
STATEMENT_B: Mary uses flash
ANSWER: No
STATEMENT_A: {Context}
STATEMENT_B: {Event}
ANSWER: 
\end{lstlisting}

%% file: prompts/llm_filtering/group/malformed-a/malformed_a-malformed_a_base.tex
\begin{lstlisting}[style=prompt,caption={Malformed-A-Base: Applies to all Context utterances.},label={prompt:malformed-a-malformed-a-base},captionpos=b]
You are an expert annotator of natural language data.
You will read a QUESTION, a STATEMENT, and some EXAMPLES.
To the best of your knowledge, and according to the examples, answer the QUESTION with yes or no.

QUESTION: Is this statement ungrammatical, misspelled, malformed, or otherwise nonsensical?
STATEMENT: Mary loses Linda arm
ANSWER: Yes
STATEMENT: Mary loses Linda's keys
ANSWER: No
STATEMENT: The mall shops back
ANSWER: Yes
STATEMENT: Mary buys many items
ANSWER: No
STATEMENT: Mary goes hopping at the mall
ANSWER: Yes
STATEMENT: Mary goes shopping at the mall
ANSWER: No
STATEMENT: {Context}
ANSWER: 
\end{lstlisting}

%% file: prompts/llm_filtering/group/malformed-a/malformed_a-malformed_a_redundant.tex
\begin{lstlisting}[style=prompt,caption={Malformed-A-Redundant: Applies to all Context utterances},label={prompt:malformed-a-malformed-a-redundant},captionpos=b]
You are an expert annotator of natural language data.
You will read a QUESTION, a STATEMENT, and some EXAMPLES.
To the best of your knowledge, and according to the examples, answer the QUESTION with yes or no.

QUESTION: Is this statement very awkward to say because it gives unnecessary information? For example, shoes always goes on feet, so it is unnecessary to mention feet if someone is putting shoes on. Similarly, lunch is always food, so it is unnecessary to mention that someone made food for lunch. People grab things with their hands, so it is unnecessary to mention the use of hands for grabbing something.
STATEMENT: Mary puts shoes on her feet
ANSWER: Yes
STATEMENT: Mary puts shoes on 
ANSWER: No
STATEMENT: Mary makes food for lunch
ANSWER: Yes
STATEMENT: Mary makes lunch
ANSWER: No
STATEMENT: Mary grabs the cup with her hand
ANSWER: Yes
STATEMENT: Mary grabs the cup
ANSWER: No
STATEMENT: {Context}
ANSWER: 
\end{lstlisting}

%% file: prompts/llm_filtering/group/malformed-ab/malformed_ab-malformed_ab_base.tex
\begin{lstlisting}[style=prompt,caption={Malformed-AB-Base: Applies to all Context-Event pairs},label={prompt:malformed-ab-malformed-ab-base},captionpos=b]
You are an expert annotator of natural language data.
You will read a QUESTION, STATEMENT_A, STATEMENT_B, and some EXAMPLES.
To the best of your knowledge, and according to the examples, answer the QUESTION with yes or no.

QUESTION: Is anything in either statement ungrammatical, misspelled, malformed, or otherwise nonsensical?
STATEMENT_A: Mary loses Linda keys
STATEMENT_B: Linda is upset
ANSWER: Yes
STATEMENT_A: Mary loses Linda's keys
STATEMENT_B: Linda is upset
ANSWER: No
STATEMENT_A: Mary loses Linda's arm
STATEMENT_B: Linda is upset
ANSWER: Yes
STATEMENT_A: Mary goes hopping at the mall
STATEMENT_B: Mary buys many ittems
ANSWER: Yes
STATEMENT_A: Mary goes shopping at the mall
STATEMENT_B: Mary buys many items
ANSWER: No
STATEMENT_A: {Context}
STATEMENT_B: {Event}
ANSWER: 
\end{lstlisting}

%% file: prompts/llm_filtering/group/malformed-ab/malformed_ab-malformed_ab_tense.tex
\begin{lstlisting}[style=prompt,caption={Malformed-AB-Tense: Applies to all Context-Event pairs},label={prompt:malformed-ab-malformed-ab-tense},captionpos=b]
You are an expert annotator of natural language data.
You will read a QUESTION, STATEMENT_A, STATEMENT_B, and some EXAMPLES.
To the best of your knowledge, and according to the examples, answer the QUESTION with yes or no.

QUESTION: Are statements A and B both in simple present tense? Note that no other tenses are included, and present progressive is NOT simple present tense.
STATEMENT_A: Mary goes shopping at the mall
STATEMENT_B: Mary was excited
ANSWER: No
STATEMENT_A: Mary goes shopping at the mall
STATEMENT_B: Mary is excited
ANSWER: Yes
STATEMENT_A: Mary goes shopping at the mall
STATEMENT_B: Mary buys many items
ANSWER: Yes
STATEMENT_A: Mary went shopping at the mall
STATEMENT_B: Mary buys many items
ANSWER: No
STATEMENT_A: Mary went shopping at the mall
STATEMENT_B: Mary is buying many items
ANSWER: No
STATEMENT_A: {Context}
STATEMENT_B: {Event}
ANSWER: 
\end{lstlisting}

%% file: prompts/llm_filtering/group/malformed-ab/malformed_ab-malformed_ab_coherent.tex
\begin{lstlisting}[style=prompt,caption={Malformed-AB-Coherent: Applies to all Context-Event pairs},label={prompt:malformed-ab-malformed-ab-coherent},captionpos=b]
You are an expert annotator of natural language data.
You will read a QUESTION, STATEMENT_A, STATEMENT_B, and some EXAMPLES.
To the best of your knowledge, and according to the examples, answer the QUESTION with yes or no.

QUESTION: Put together, are statements A and B coherent? In other words, if statements A and B make sense and are related to each other, answer Yes. Otherwise, answer No.
STATEMENT_A: Mary goes shopping at the mall
STATEMENT_B: Mary buys many items
ANSWER: Yes
STATEMENT_A: Mary goes shopping at the mall
STATEMENT_B: The mall shops back
ANSWER: No
STATEMENT_A: Mary goes shopping at the mall
STATEMENT_B: The elf smiles back
ANSWER: No
STATEMENT_A: {Context}
STATEMENT_B: {Event}
ANSWER: 
\end{lstlisting}

%% file: prompts/llm_filtering/indiv/brevity.tex
\begin{lstlisting}[style=prompt,caption={Brevity: Applies to all Context-Event pairs},label={prompt:brevity},captionpos=b]
You are an expert annotator of natural language data.
You will read a QUESTION, STATEMENT_A, STATEMENT_B, and some EXAMPLES.
To the best of your knowledge, and according to the examples, answer the QUESTION with yes or no.

QUESTION: Could statement B be edited down to be more brief, stripped of descriptive modifiers, while still retaining its meaning? If minor modifiers or details can be deleted, answer Yes. Otherwise, answer No. Please be very careful. If you think the entirety of statement B can be deleted, answer No.
STATEMENT_A: Mary purchases the bike
STATEMENT_B: Mary inspects it briefly
ANSWER: Yes
STATEMENT_A: Mary purchases the bike
STATEMENT_B: Mary inspects it
ANSWER: No
STATEMENT_A: Mary watches television
STATEMENT_B: Mary turns it off later
ANSWER: Yes
STATEMENT_A: Mary watches television
STATEMENT_B: Mary turns it off
ANSWER: No
STATEMENT_A: Mary takes a shopping cart to the store
STATEMENT_B: Mary fills the cart with products
ANSWER: Yes
STATEMENT_A: Mary takes a shopping cart to the store
STATEMENT_B: Mary fills the cart with cheese
ANSWER: No
STATEMENT_A: {Context}
STATEMENT_B: {Event}
ANSWER: STATEMENT_A: Context
STATEMENT_B: Event
ANSWER: 
\end{lstlisting}

%% file: prompts/llm_filtering/indiv/definite.tex
\begin{lstlisting}[style=prompt,caption={Definite: Applies to only subaction Events.},label={prompt:definite},captionpos=b]
You are an expert annotator of natural language data.
You will read a QUESTION, STATEMENT_A, STATEMENT_B, and some EXAMPLES.
To the best of your knowledge, and according to the examples, answer the QUESTION with yes or no.

QUESTION: If statement B has an object with 'a', 'an', or a possessive at preceding it, can it be replaced with 'the' or one of 'her/his/their/its' and still make sense? If statement B already has an object modified by 'the', answer No. If statement B no article or possessive, or no object, answer No Only answer yes if statement B has 'a', 'an', or a possessive modifying the object, AND it can be replaced with the while still making sense.
STATEMENT_A: Mary buys new sheets
STATEMENT_B: Mary selects a color
ANSWER: Yes
STATEMENT_A: Mary buys new sheets
STATEMENT_B: Mary selects the color
ANSWER: No
STATEMENT_A: Mary goes bowling with Linda
STATEMENT_B: Mary picks a bowling ball
ANSWER: Yes
STATEMENT_A: Mary goes bowling with Linda
STATEMENT_B: Mary picks her bowling ball
ANSWER: No
STATEMENT_A: Mary makes friendship bracelets
STATEMENT_B: Mary ties knots
ANSWER: Yes
STATEMENT_A: Mary makes friendship bracelets
STATEMENT_B: Mary ties the knots
ANSWER: No
STATEMENT_A: {Context}
STATEMENT_B: {Event}
ANSWER: 
\end{lstlisting}

%% file: prompts/llm_filtering/indiv/narrative.tex
\begin{lstlisting}[style=prompt,caption={Narration: Applies to all Context-Event pairs},label={prompt:narrative},captionpos=b]
You are an expert annotator of natural language data.
You will read a QUESTION, STATEMENT_A, STATEMENT_B, and some EXAMPLES.
To the best of your knowledge, and according to the examples, answer the QUESTION with yes or no.

QUESTION: Paired together, do statements A and B seem like they would be found in a novel as opposed to a text message or an online forum? Clues to look for: - Verbs that are loaded, colorful, or literary: 'tries' to clean up a spill is loaded because it implies some failure to complete the task. 'Rushes' instead of 'goes' or 'clutches' instead of 'uses' are more colorful/literary than their alternatives. - Descriptions of minute actions: e.g. 'Mary turns her head' - Information, like an internal monologue, that a regular friend would not be privy too, but that an author of a story might write.
STATEMENT_A: Mary buys new sheets
STATEMENT_B: Mary thinks about her bed
ANSWER: Yes
STATEMENT_A: Mary buys new sheets
STATEMENT_B: Mary washes them
ANSWER: No
STATEMENT_A: Mary spills a drink
STATEMENT_B: Mary tries to clean it up
ANSWER: Yes
STATEMENT_A: Mary spills a drink
STATEMENT_B: Mary cleans it up
ANSWER: No
STATEMENT_A: Mary hears a thud
STATEMENT_B: Mary turns her head
ANSWER: Yes
STATEMENT_A: Mary hears a thud
STATEMENT_B: Mary gets up
ANSWER: No
STATEMENT_A: Mary cuts Linda's hair short
STATEMENT_B: Linda looks in the mirror
ANSWER: Yes
STATEMENT_A: Mary cuts Linda's hair short
STATEMENT_B: There is a mirror
ANSWER: No
STATEMENT_A: Mary roasts marshmallows
STATEMENT_B: Mary clutches a stick
ANSWER: Yes
STATEMENT_A: Mary roasts marshmallows
STATEMENT_B: Mary uses a stick
ANSWER: No
STATEMENT_A: {Context}
STATEMENT_B: {Event}
ANSWER: 
\end{lstlisting}

%% file: prompts/llm_filtering/group/life/life-dailylife.tex
\begin{lstlisting}[style=prompt,caption={Life-Daily-Life: Applies to all Context utterances.},label={prompt:life-dailylife},captionpos=b]
You are an expert annotator of natural language data.
You will read a QUESTION, a STATEMENT, and some EXAMPLES.
To the best of your knowledge, and according to the examples, answer the QUESTION with yes or no.

QUESTION: Does this statement describe a situation or activity which can be considered abnormal or hyperspecific? For example, fishing as a general activity is normal, but adding certain details to the statement can make it either abnormal or hyperspecific. Murder or other kinds of non-petty violence do happen in the world daily, but these are not considered a part of normal life.
STATEMENT: Mary goes fishing
ANSWER: No
STATEMENT: Mary goes fishing for treasure
ANSWER: Yes
STATEMENT: Mary brings four rods to the lake to fish
ANSWER: Yes
STATEMENT: Mary makes dinner for Linda
ANSWER: No
STATEMENT: Mary gives Linda a present
ANSWER: No
STATEMENT: Mary gives Linda poison
ANSWER: Yes
STATEMENT: Mary meets with her dad
ANSWER: No
STATEMENT: Mary meets with her dad's fortune teller
ANSWER: Yes
STATEMENT: Mary finds out her dad is not her biological dad
ANSWER: Yes
STATEMENT: {Context}
ANSWER: 
\end{lstlisting}

%% file: prompts/llm_filtering/group/life/life-lifetime.tex
\begin{lstlisting}[style=prompt,caption={Life-Lifetime: Applies to all Context utterances.},label={prompt:life-lifetime},captionpos=b]
You are an expert annotator of natural language data.
You will read a QUESTION, a STATEMENT, and some EXAMPLES.
To the best of your knowledge, and according to the examples, answer the QUESTION with yes or no.

QUESTION: Is this statement about something that happens as a milestone in a person's lifetime? Note that if it can be a milestone, but it can also be a regular occurrence for certain people's jobs or hobbies, your answer should be no. An example of this is winning a competition: this can happen regularly for athletes and other types of people. But something like graduating high school, by definition, can only happen once in a lifetime. And giving birth can happen a few times, but is always a significant event.
STATEMENT: Mary attends her class
ANSWER: No
STATEMENT: Mary graduates high school
ANSWER: Yes
STATEMENT: Mary gives Linda a present
ANSWER: No
STATEMENT: Mary gives birth
ANSWER: Yes
STATEMENT: {Context}
ANSWER: 
\end{lstlisting}

%% file: prompts/llm_filtering/indiv/gap.tex
\begin{lstlisting}[style=prompt,caption={Gap: Applies to all next-action Events.},label={prompt:gap},captionpos=b]
You are an expert annotator of natural language data.
You will read a QUESTION, STATEMENT_A, STATEMENT_B, and some EXAMPLES.
To the best of your knowledge, and according to the examples, answer the QUESTION with yes or no.

QUESTION: Is there a very wide time-gap or conceptual gap between statements A and B?
STATEMENT_A: Mary waters Linda's plants
STATEMENT_B: Linda's plants stay healthy
ANSWER: Yes
STATEMENT_A: Mary waters Linda's plants
STATEMENT_B: Linda thanks her
ANSWER: No
STATEMENT_A: Mary finds a wife
STATEMENT_B: Mary plans a wedding
ANSWER: Yes
STATEMENT_A: Mary gets engaged
STATEMENT_B: Mary plans a wedding
ANSWER: No
STATEMENT_A: Mary walks to the library
STATEMENT_B: Mary reads a book
ANSWER: Yes
STATEMENT_A: Mary walks to the library
STATEMENT_B: Mary checks out a book
ANSWER: No
STATEMENT_A: {Context}
STATEMENT_B: {Event}
ANSWER: 
\end{lstlisting}

%% file: prompts/llm_filtering/indiv/scope.tex
\begin{lstlisting}[style=prompt,caption={Scope: Applies to just subaction Events.},label={prompt:scope},captionpos=b]
You are an expert annotator of natural language data.
You will read a QUESTION, STATEMENT_A, STATEMENT_B, and some EXAMPLES.
To the best of your knowledge, and according to the examples, answer the QUESTION with yes or no.

QUESTION: Can statement A be considered much wider in scope than statement B?
STATEMENT_A: Mary takes many photos
STATEMENT_B: Mary adjusts the lens
ANSWER: Yes
STATEMENT_A: Mary takes a photo
STATEMENT_B: Mary adjusts the lens
ANSWER: No
STATEMENT_A: Mary plays chess
STATEMENT_B: Mary plans her next move
ANSWER: Yes
STATEMENT_A: Mary moves her chess piece
STATEMENT_B: Mary plans her next move
ANSWER: No
STATEMENT_A: Mary sails along the coast
STATEMENT_B: Mary adjusts the sail
ANSWER: Yes
STATEMENT_A: Mary sails along the coast
STATEMENT_B: Mary documents it
ANSWER: No
STATEMENT_A: Mary spends the day at the beach
STATEMENT_B: Mary sets up a towel
ANSWER: Yes
STATEMENT_A: Mary spends the day at the beach
STATEMENT_B: Mary swims
ANSWER: No
STATEMENT_A: {Context}
STATEMENT_B: {Event}
ANSWER: 
\end{lstlisting}

%% file: prompts/llm_filtering/indiv/passive.tex
\begin{lstlisting}[style=prompt,caption={Passive: Applies to all Context utterances.},label={prompt:passive},captionpos=b]
You are an expert annotator of natural language data.
You will read a QUESTION, a STATEMENT, and some EXAMPLES.
To the best of your knowledge, and according to the examples, answer the QUESTION with yes or no.

QUESTION: Does this statement contain passive voice?
STATEMENT: Mary gets told off by Linda
ANSWER: Yes
STATEMENT: Linda tells Mary off
ANSWER: No
STATEMENT: {Context}
ANSWER: 
\end{lstlisting}

%% file: prompts/llm_filtering/indiv/standalone.tex
\begin{lstlisting}[style=prompt,caption={Standalone: Applies to all Context utterances.},label={prompt:standalone},captionpos=b]
You are an expert annotator of natural language data.
You will read a QUESTION, a STATEMENT, and some EXAMPLES.
To the best of your knowledge, and according to the examples, answer the QUESTION with yes or no.

QUESTION: Can this statement stand alone without extra context? Answer Yes if the statement makes no references to a previous context and it makes sense without extra context.
STATEMENT: Mary leaves the shop
ANSWER: Yes
STATEMENT: Mary goes back to the shop
ANSWER: No
STATEMENT: Mary crashes into a car
ANSWER: Yes
STATEMENT: Mary crashes into another car
ANSWER: No
STATEMENT: Mary gives Linda money
ANSWER: Yes
STATEMENT: Mary gets Linda's money back
ANSWER: No
STATEMENT: Mary gets Linda's money
ANSWER: Yes
STATEMENT: Mary loses her keys
ANSWER: Yes
STATEMENT: Mary finds the keys at home
ANSWER: No
STATEMENT: {Context}
ANSWER: 
\end{lstlisting}

%% file: prompts/llm_filtering/indiv/vague.tex
\begin{lstlisting}[style=prompt,caption={Vague: Applies to all Context-Event pairs},label={prompt:vague},captionpos=b]
You are an expert annotator of natural language data.
You will read a QUESTION, STATEMENT_A, STATEMENT_B, and some EXAMPLES.
To the best of your knowledge, and according to the examples, answer the QUESTION with yes or no.

QUESTION: Put together, are statements A and B extremely underspecified? Note that if the statements are general, but not extremely underspecified, your answer should be No. Your answer should only be Yes if the generalness exceeds a threshold where a core part of the meaning of the text seems to be missing.
STATEMENT_A: Mary sees the look on Linda's face
STATEMENT_B: Mary reacts
ANSWER: Yes
STATEMENT_A: Mary sees the angry look on Linda's face
STATEMENT_B: Mary reacts
ANSWER: No
STATEMENT_A: Mary sees the look on Linda's face
STATEMENT_B: Mary calls it out
ANSWER: No
STATEMENT_A: Mary starts a club
STATEMENT_B: Mary is interested in a topic
ANSWER: Yes
STATEMENT_A: Mary starts a chess club
STATEMENT_B: Mary is interested in chess
ANSWER: No
STATEMENT_A: {Context}
STATEMENT_B: {Event}
ANSWER: 
\end{lstlisting}

%% file: prompts/llm_filtering/indiv/typicality_q.tex
\begin{lstlisting}[style=prompt,caption={No-Typicality: Applies to all Context-Event pairs},label={prompt:typicality-q},captionpos=b]
You are an expert annotator of natural language data.
You will read a QUESTION, a TEXT, and some EXAMPLES.
To the best of your knowledge, and according to the examples, answer the QUESTION with yes or no.

QUESTION: Does the following question "When {}, how often is {}?" make sense grammatically and meaning-wise? For example, if statement B is about an inherent trait of someone, the question does not make sense and your answer should be No. Generally, answer No if it does not make sense to think about a situational relationship between statements A and B. Otherwise, answer Yes.
TEXT: When Mary finds a stray dog, how often does she pet it?
ANSWER: Yes
TEXT: When Mary finds a stray dog, how often is she an animal lover?
ANSWER: No
TEXT: When Mary builds a house, how often does she take a whole year?
ANSWER: Yes
TEXT: When Mary builds Linda's house, how often does Linda move in?
ANSWER: No
TEXT: When Mary makes friendship bracelets, how often does she use more than one bead shape?
ANSWER: Yes
TEXT: When Mary makes friendship bracelets, how often does she enjoy crafting as a hobby?
ANSWER: No
TEXT: When Mary sees shoes for sale, how often does she buy them?
ANSWER: Yes
TEXT: When Mary sees shoes for sale, how often is she interested in fashion?
ANSWER: No
TEXT: When {}, how often is {}?
ANSWER: 
\end{lstlisting}

%% file: prompts/llm_filtering/indiv/explain.tex
\begin{lstlisting}[style=prompt,caption={Explain: Applies to only state Events.},label={prompt:explain},captionpos=b]
You are an expert annotator of natural language data.
You will read a QUESTION, STATEMENT_A, STATEMENT_B, and some EXAMPLES.
To the best of your knowledge, and according to the examples, answer the QUESTION with yes or no.

QUESTION: Is statement B the answer for why statement A happened? In other words, can you easily infer the word BECAUSE between statements A and B? "Context BECAUSE Event" Please consider your answer in terms of coherence relations. In this case, is BECAUSE more likely than AS A RESULT? If AS A RESULT is more likely than or equally likely to BECAUSE, answer No. If statement B is just background information about the scenario, answer No.
STATEMENT_A: Mary takes a shopping cart to the store
STATEMENT_B: Mary plans to buy multiple items
ANSWER: Yes
STATEMENT_A: Mary takes a shopping cart to the store
STATEMENT_B: Mary has a list
ANSWER: No
STATEMENT_A: Mary goes back to sleep
STATEMENT_B: Mary is tired
ANSWER: Yes
STATEMENT_A: Mary goes back to sleep
STATEMENT_B: Mary is on a bed
ANSWER: No
STATEMENT_A: Mary covers her head
STATEMENT_B: Mary feels cold
ANSWER: Yes
STATEMENT_A: Mary covers her head
STATEMENT_B: Mary is outside
ANSWER: No
STATEMENT_A: {Context}
STATEMENT_B: {Event}
ANSWER: 
\end{lstlisting}

%% file: prompts/llm_filtering/group/result/result-result_ab.tex
\begin{lstlisting}[style=prompt,caption={Result-AB: Applies to only state Events.},label={prompt:result-result-ab},captionpos=b]
You are an expert annotator of natural language data.
You will read a QUESTION, STATEMENT_A, STATEMENT_B, and some EXAMPLES.
To the best of your knowledge, and according to the examples, answer the QUESTION with yes or no.

QUESTION: Is statement B the direct result of statement A happening? In other words, is the text meaning equivalent to "Event, AND AS A RESULT Context"?
STATEMENT_A: Mary smells smoke
STATEMENT_B: Mary is concerned
ANSWER: Yes
STATEMENT_A: Mary smells smoke
STATEMENT_B: Mary has something in the oven
ANSWER: No
STATEMENT_A: Mary wins the prize
STATEMENT_B: Mary is overjoyed
ANSWER: Yes
STATEMENT_A: Mary wins the prize
STATEMENT_B: Mary is far ahead of everyone else
ANSWER: No
STATEMENT_A: {Context}
STATEMENT_B: {Event}
ANSWER: 
\end{lstlisting}

%% file: prompts/llm_filtering/group/result/result-result_ba.tex
\begin{lstlisting}[style=prompt,caption={Result-BA: Applies to only state Events.},label={prompt:result-result-ba},captionpos=b]
You are an expert annotator of natural language data.
You will read a QUESTION, STATEMENT_A, STATEMENT_B, and some EXAMPLES.
To the best of your knowledge, and according to the examples, answer the QUESTION with yes or no.

QUESTION: Is statement A the direct result of statement B happening? In other words, is the text meaning equivalent to "Context, AND AS A RESULT Event"?
STATEMENT_A: Mary is concerned
STATEMENT_B: Mary smells smoke
ANSWER: Yes
STATEMENT_A: Mary has something in the oven
STATEMENT_B: Mary smells smoke
ANSWER: No
STATEMENT_A: Mary is overjoyed
STATEMENT_B: Mary wins the prize
ANSWER: Yes
STATEMENT_A: Mary is far ahead of everyone else
STATEMENT_B: Mary wins the prize
ANSWER: No
STATEMENT_A: {Context}
STATEMENT_B: {Event}
ANSWER: 
\end{lstlisting}

%% file: prompts/llm_filtering/indiv/habitual.tex
\begin{lstlisting}[style=prompt,caption={Habitual: Applies to all Context utterances.},label={prompt:habitual},captionpos=b]
You are an expert annotator of natural language data.
You will read a QUESTION, a STATEMENT, and some EXAMPLES.
To the best of your knowledge, and according to the examples, answer the QUESTION with yes or no.

QUESTION: Is this a statement explicitly about somebody's habits?
STATEMENT: Mary plays football weekly
ANSWER: Yes
STATEMENT: Mary plays in the football game
ANSWER: No
STATEMENT: {Context}
ANSWER: 
\end{lstlisting}